\pdfoutput=1 
\documentclass[10pt,journal,compsoc]{IEEEtran}
\usepackage[T1]{fontenc}
\usepackage{cite}

\usepackage{subcaption}

\usepackage{blkarray}
\usepackage{array}
\usepackage{lmodern}
\usepackage{graphicx}
\usepackage{algorithm}
\usepackage{algpseudocode}
\usepackage{bm}
\usepackage{url}
\usepackage{float}
\usepackage{multirow}
\usepackage{amssymb,amsmath,amsfonts,amsthm}
\usepackage{ifxetex,ifluatex}

\DeclareMathOperator*{\diag}{diag}
\DeclareMathOperator*{\argmax}{arg\,max}

\DeclareMathOperator*{\OptSpace}{OptSpace}
\DeclareMathOperator*{\NMF}{NMF}

\DeclareMathOperator*{\MLqE}{MLqE}
\DeclareMathOperator*{\MAP}{MAP}
\DeclareMathOperator*{\AIC}{AIC}
\DeclareMathOperator*{\BIC}{BIC}

\DeclareMathOperator*{\MSE}{MSE}

\usepackage[english]{babel}
\usepackage{graphicx}

\usepackage[utf8]{inputenc}
\usepackage{bm}

\newtheorem{my.def}{Definition}
\newtheorem{my.thm}{Theorem}
\newtheorem{my.lemma}{Lemma}
\newtheorem{my.assumption}{Assumption}

\DeclareMathOperator{\pdiag}{diag}

\makeatletter
\renewcommand*\env@matrix[1][*\c@MaxMatrixCols c]{%
  \hskip -\arraycolsep
  \let\@ifnextchar\new@ifnextchar
  \array{#1}}
\makeatother

\begin{document}

\title{A model selection approach for clustering a multinomial sequence with non-negative factorization}

\author{Nam~H.~Lee,
        Runze~Tang,
	Carey~E.~Priebe,
        Michael~Rosen%
        \IEEEcompsocitemizethanks{
        \IEEEcompsocthanksitem 
        Nam Lee, Carey Priebe and Runze Tang
        are with Department of Applied 
        Mathematics and Statistics, Johns 
        Hopkins University\protect\\
        nhlee@jhu.edu
        \IEEEcompsocthanksitem
        Michael Rosen is with 
		Armstrong Institute for Patient 
        Safety and Quality, Johns Hopkins 
        University        
        }%
        }

\IEEEcompsoctitleabstractindextext{
\begin{abstract} 
We consider a problem of clustering a sequence of multinomial observations by way of a model selection criterion.  We propose a form of a penalty term for the model selection procedure.  Our approach subsumes both the conventional AIC and BIC criteria but also extends the conventional criteria in a way that 
it can be applicable also to a sequence of sparse multinomial observations, where even within a same cluster, the number of multinomial trials may be different for different observations.   In addition, as a preliminary estimation step to maximum likelihood estimation, and more generally, to maximum $L_{q}$ estimation, we propose to use reduced rank projection in combination with non-negative factorization.  
We motivate our approach by showing that our model selection criterion and preliminary estimation step yield consistent estimates under simplifying assumptions.  
We also illustrate our approach through numerical experiments using real and simulated data.
\end{abstract}

}

\maketitle

\section{Introduction}

We consider a problem of clustering a sequence of multinomial observations.
To be specific, consider a sequence $(X_{1}, X_{2}, \ldots, X_{T})$ of independent multinomial 
random vectors taking values in $\mathbb N^{d}$ for some $d \gg T$, where $\mathbb N = \{0,1,2,\ldots, \}$.
For each $t=1,\ldots, T$, each $X_{t}$ is a multinomial random vector such that the number of trials is $N_{t}$ and 
the success probability vector is $P_{t}$.   To simplify our notation, we write $X_{t} \sim {MN}(N_{t}, P_{t})$.
We allow the value of $N_{t}$ to depend on the value of $t$ and similarly, we allow the value of 
$P_{t}$ to depend on the value of $t$.  Moreover, to formulate our clustering problem, we assume that 
there is a finite collection $\mathcal Q = \{Q_{1}, Q_{2},\ldots, Q_{K}\}$  of $d$-dimensional probability vectors 
such that $\{P_{1},P_{2},\ldots, P_{T}\} = \mathcal Q$.  Since $\{P_{1},P_{2},\ldots, P_{T}\} = \mathcal Q$, it follows that
 for each $k=1,\ldots, K$, there exists $t=1,\ldots, T$ such that 
$P_{t} = Q_{k}$.  For each $t$, we let $\kappa(t) = k$ provided that 
$P_{t} = Q_{k}$, and to simplify our notation, we may also write $t \in k(\kappa)$ to mean $\kappa(t) = k$.  

We assume that the value of $K$, $\kappa$ and $\mathcal Q$ are unknown, but
the value of $(X_{1},\ldots, X_{T})$ is observed and forms the basis for statistical inference.  
Let $\theta = (\kappa, \mathcal Q)$, and let $\Theta(K)$ be the set of all possible values for $\theta$.  
Note that $\kappa$ can be represented with an element in $\{1,\ldots, K\}^{T}$ and $\mathcal Q$ can be 
represented with an element in $[0,1]^{d\times K}$, i.e., a $d\times K$ non-negative matrix, where each column 
sums to $1$. Since the value of $K$ is assumed to be unknown, from a parameter estimation point of view, one also must consider the set $\Theta(K)$ for all $K=1,2,\ldots, T$ as a potential set to which the true parameter $\theta$ belongs.  
Then, we let 
\begin{align}
\Theta = \cup_{K=1}^{T} \Theta(K).
\end{align}

Henceforth, we write $K^{*}$ and $\theta^{*} = (\kappa^{*},\mathcal Q^{*}) $ for the parameter that generates the data 
$(X_{1},X_{2},\ldots, X_{T})$.  The estimates of 
$K$, $\kappa$ and $\mathcal Q$ are denoted by 
$\widehat K$, $\widehat \kappa$ and $\widehat{ \mathcal Q}$ respectively, 
and we now use the letters $K$, $\kappa$, $\mathcal Q$ for a generic value that $\widehat K, \widehat \kappa, \widehat{\mathcal Q}$ can take respectively.   

In this paper, we propose to take $\widehat K$, $\widehat \kappa$ and $\widehat{\mathcal Q}$ to be solutions to 
the optimization problems specified in \eqref{eqn:partialnegloglik} and \eqref{eqn:base-form-of-Delta(K)}.  
To this end, the rest of this paper is organized as follows.  

In Section 2.1, we present the overall description of our 
approach, specifically introducing \eqref{eqn:partialnegloglik} and \eqref{eqn:base-form-of-Delta(K)}.   
In Section 2.2, we present a preliminary estimation technique, which can be used prior to  
performing a numerical search for the solution to \eqref{eqn:partialnegloglik}.  
In Section 2.3, we specify the penalty term for our model selection criteria in \eqref{eqn:base-form-of-Delta(K)}.    

In Section 3.1, we motivate our choice of the penalty term in \eqref{eqn:penterm-aicc} through Theorem \ref{thm:result/main} and \ref{thm:main:ModelSelectThm}.
In Section 3.2, we motivate, in Theorem \ref{thm:meantotalinteraction}, our usage of the reduced rank projection step within our estimation steps. 

In Section \ref{sec:NumericalResult}, we compare our model selection criterion with the conventional AIC via a Monte Carlo simulation experiment.  We also study, through our approach, 
a two-sample test problem for comparing two graphs.  This is done using simulated data sets, 
as well as a real data set involving the chemical and electrical connectivity structure of neurons of a C.~elegan.  
Lastly, we apply our technique to a problem of determining the model dimension associated 
with the so-called Swimmer dataset which is well known to the non-negative factorization community (c.f.~\cite{Donoho03whendoes})

\section{Background materials}
\subsection{General framework}
To begin, we represent the sequence $(X_{1}, X_{2}, \ldots, X_{T})$ as an integer-valued random matrix $X$ so that 
the element in the $t$th column of $X$ is $X_{t}$.  With slight abuse of notation, we denote the $i$th row and the $t$th column of $X$ 
by $X_{it}$.  Since the sample value of $X$ is known, it follows that the sample values of $N_{1},N_{2},\ldots, N_{T}$ are known.  Then, 
it follows that, denoting by $P$ the $d\times T$ matrix whose $t$th column is given by $P_{t}$, we have
\begin{align}
\mathbf E_{\theta}[X \diag(N_{1},\ldots, N_{T})^{-1}]  = P, \label{eqn:NMFdecomposition000}
\end{align}
where $\diag(N_{1},\ldots, N_{T})$ is a $T\times T$ diagonal matrix such that its $t$th diagonal is $N_{t}$ and the expectation is taken 
with respect to the probability measure specified by $\theta$.  
Moreover, in general, it can be seen that $P$ can be factored as a product of two column stochastic non-negative matrices, 
namely, $W$ and $H$.  Specifically,
\begin{align}
P = W H,  \label{eqn:NMFdecomposition111}
\end{align}
where for each $k=1,\ldots, K$, the $k$th column $W_{k}$ of $W$ is $Q_{k}$ and for each 
$t = 1,\ldots, T$, the $t$th column $H_{t}$ of $H$ is the basis vector of $\mathbb R^{K}$ such that 
its $k$-th entry is $1$ if and only if $\kappa(t) = k$.  

In light of \eqref{eqn:NMFdecomposition000}
and \eqref{eqn:NMFdecomposition111},  when $X$ is observed  without noise, i.e., $\mathbf E[X] = X$, and given that the value of $K$ is known, application of a non-negative factorization algorithm can recover $W$ and $H$ from $P$.  However, in general, $X$ is random.  
Specifically, we have that for each $\theta \in \Theta$, 
\begin{align*}
f_{X}(X_{1},X_{2},\ldots, X_{T}| \theta) 
= \prod_{t=1}^{T} \left({N_{t} \choose X_{t}}  \prod_{i=1}^{d}P_{it}^{X_{it}}\right),
\end{align*}
where for simplicity, we write
\begin{align*}
{ N_{t} \choose X_{t} } ={ N_{t} \choose X_{1t}, X_{2t}, \ldots, X_{dt} }.
\end{align*}
Alternatively, we may also write, by grouping according to the value of $\kappa(t)$, that
\begin{align*}
f_{X}(X_{1},X_{2},\ldots, X_{T}| \theta) 
=  \left(  \prod_{k=1}^{K} \prod_{i=1}^{d}Q_{ik}^{\sum_{t \in k(\kappa)}X_{it}}\right) \prod_{t=1}^{T} {N_{t} \choose X_{t}}.
\end{align*}
For simplicity, we may write  
\begin{align*}
N= \sum_{t=1}^{T} N_{t}.
\end{align*}

Then, for each $K = 1,\ldots, T$, let $\widehat \theta(K)$ be an maximum $L_{q}$ estimate of 
$\theta^{*}$ with the restriction that the value of $\widehat \theta(K)$ must be an element of $\Theta(K)$, for 
some value of $q$ (c.f.~\cite{ferrari2010}).  
Specifically, for each $K$ and $q$,  we denote by $\widehat \theta(K;q)$, an maximum $L_{q}$ estimate of $\theta^{*}$ given $K$, and we have
\begin{align}
\widehat \theta(K;q) \in \argmax_{\theta \in \Theta(K)} 
\sum_{t=1}^{T} \sum_{i=1}^{d} X_{i}(t) \left(\frac{Q_{i,\kappa(t)}^{1-q} - 1 }{1-q}\right).  \label{eqn:partialnegloglik}
\end{align}
Note that by taking $q < 1$ to $1$ in limit, then we see that 
\begin{align*}
\quad & \lim_{q\rightarrow1}\sum_{t=1}^{T} \sum_{i=1}^{d} X_{i}(t) \left(\frac{Q_{i,\kappa(t)}^{1-q} - 1 }{1-q}\right) \\
= & \sum_{t=1}^{T} \sum_{i=1}^{d} X_{i}(t) \log(Q_{i,\kappa(t)}),
\end{align*}
and as such, $\widehat \theta(K;1)$ reduces to a maximum likelihood estimate.
When the value of $q$ is clear from context, we suppress $q$ from $\widehat \theta(K;q)$ and write 
$\widehat \theta(K)$ instead.

Then, we let $\widehat K$ be the smallest values of $K$ that minimizes the value of the following expression:
\begin{align}
\Delta(K) := 
 \sum_{t=1}^{T} D_{t}(X_{t}, \widehat Q_{\widehat \kappa(t)}) + \texttt{penalty}(K), \label{eqn:base-form-of-Delta(K)}
\end{align}
where  $\texttt{penalty}(K)$ is assumed to be chosen a priori, $\widetilde P_{it} = X_{it}/N_{t}$, and 
\begin{align*}
\quad D_{t}(X_{t}, \widehat Q_{\widehat \kappa(t)})
& := D_{KL}({\widetilde P_{t}} || \widehat Q_{\widehat \kappa(t)}) - \sum_{i=1}^{d}  {\widetilde P_{it}}  \log( {\widetilde P_{it}} ) \\
& = - \sum_{i=1}^{d}  {\widetilde P_{it}} \log(\widehat Q_{i \widehat \kappa(t)}),
\end{align*}
denoting by $D_{KL}(\mu_{1}||\mu_{2})$ the Kullback-Leibler divergence of $\mu_{2}$ from $\mu_{1}$.  For reference, we let 
\begin{align*}
D(\widehat{\mathcal Q},  \widehat \kappa)  
:=
D(\widehat{\mathcal Q},  \widehat \kappa ; X)  
:= \sum_{t=1}^{T} D_{t}(X_{t}, \widehat Q_{\widehat \kappa(t)}).
\end{align*}

\subsection{Preliminary estimation prior to $\MLqE$}
For our model selection problem, for each $K$, an estimator $\widehat P$ of $P^{*}$ as function of $\widetilde P$ 
must minimize the size $\|\widehat P - P^{*}\|$ of error while also allowing for non-negative factorization, i.e., 
$\widehat P = \widehat W \widehat H$ where $\widehat W$ and $\widehat H$ are $d\times K$ and 
$K\times T$ non-negative matrices.  Directly computing an $\MLqE$ to achieve this can be done numerically
with varying degree of complexity, but in all cases, starting the search for $\MLqE$ near $P^{*} = W^{*} H^{*}$ can be beneficial. 
To achieve this approximately for initializing our $\MLqE$ search algorithm
for numerical experiments, we propose a multi-step procedure in which $\OptSpace$ and $\NMF$ are used together.  For more details on $\OptSpace$ (and also on
USVT, a related approach), we direct the reader to \cite{kmo10noise} (and \cite{Chatterjee2013}), and for $\NMF$, to 
\cite{HKimHPark2008}, \cite{berry2007algorithms} and \cite{Brunet23032004}.

Iteratively searching for a solution to the estimation problem in \eqref{eqn:partialnegloglik} can be computationally expensive. 
An approximate solution, which can be used as the initial point of the search, can be obtained in four steps, which are listed 
in Algorithm \ref{algo:mainalgorithm} collectively for convenience. 

First we take a reduced rank projection of $X$ at the rank $K$.   
Specifically, we first compute the singular value decomposition of $X = U \Sigma V^{\top}$ with the diagonal of $\Sigma$ being sorted in 
a decreasing order, e.g. $\Sigma_{11} \ge \Sigma_{22}$.
Then, we take 
\begin{align}
\widehat X = \OptSpace(X;K) := \widehat U \widehat \Sigma \widehat V^{\top},
\end{align}
where $\widehat \Sigma$ is the upper $K\times K$ block of $\Sigma$,
and $\widehat U$ and $\widehat V$ are the first $K$ columns of $U$ and $V$ respectively.  

Because $\widehat X$ need not be non-negative, we then reset the negative entries of $\widehat X$ to zero.
However, the resetting the negative entries of $\widehat X$ to zero can change the rank of $\widehat X$.

To correct this, after computing $\widehat P = \widehat X\pdiag(\bm 1^\top \widehat X)^{-1}$,
we further perform non-negative factorization, which means we minimize $\|\widehat P -  W  H\|_F$
by running over all possible pairs of $d\times K$ matrix $W \ge 0$ and $K\times T$ matrix $H \ge 0$ (c.f.~\cite{HKimHPark2008}).  

In the last step, we define $\widehat \kappa$ by letting, for each $t =1,\ldots, T$, $\widehat \kappa(t) = k$ if and only if $\widehat H_{kt} \ge \widehat H_{k^{\prime}t}$
for all $k^{\prime} = 1,\ldots, K$, where a tie, if any exists, is resolved by uniformly choosing among the tied indices. 
Then, estimate $\widehat{\mathcal Q} = \{\widehat W_{1},\ldots, \widehat W_{K}\}$.

The aforementioned steps for obtaining $\widehat \kappa$ and $\widehat Q$ are collectively denoted as $\MAP(\widehat W,\widehat H)$ in the listing of 
Algorithm \ref{algo:mainalgorithm}.

Upon obtaining the initial value of $(\widehat Q, \widehat \kappa)$, one can perform a numerical iterative search for $\MLqE$, for example, using a variational method, an EM algorithm, an MCMC method, or a brute force iterative search.  
For an interested reader, in Section \ref{sec:MCMC-MLqE-objective}, we outline an objective function to be maximized for an MCMC approach.  In all cases, it is known that a good initialization of the chosen algorithm can improve its rate of convergence as well as allowing the algorithm to avoid a local stationary point.
On the other hand, the number $K$ of clusters to be estimated still needs to be supplied, for each of these algorithms.

\begin{algorithm}[t!]
    \caption{Preliminary Estimation prior to  $\MLqE$}
    \label{algo:mainalgorithm}
\begin{algorithmic}[1]
    \Require $K=1,\ldots, T$ and data matrix $X$
    \vskip0.1in
    \Procedure {Preliminary Estimation}{}
    \State $\widehat X \leftarrow \OptSpace(X;K)$
    \State $\widehat P \leftarrow \widehat X \pdiag(\bm 1^\top \widehat X)^{-1}$    
    \State $(\widehat W,\widehat H) \leftarrow \NMF(\widehat P;K)$
    \State $(\widehat \kappa, \widehat{\mathcal Q}) \leftarrow \MAP(\widehat W, \widehat H)$

    \State \Return $(\widehat \kappa, \widehat{\mathcal Q})$.
    \EndProcedure
\end{algorithmic}
\end{algorithm}

\subsection{Model selection criterion}\label{sec:ModelSelection}

For many application, the following standard model selection criteria are often used:
\begin{align}
\Delta_{\AIC}(K) = -\log(f_{X}(X | \widehat \theta(K))) + \texttt{penalty}_{\AIC}(K), \label{eqn:DeltaAIC-classic}\\
\Delta_{\BIC}(K) = -\log(f_{X}(X | \widehat \theta(K))) + \texttt{penalty}_{\BIC}(K), \label{eqn:DeltaBIC-classic}
\end{align}
where  

$\texttt{penalty}_{\AIC}(K) := (d-1)K$, 
$\texttt{penalty}_{\BIC}(K) := (d-1)K \log\left(N\right)$,

and $\widehat \theta(K)$ is chosen to be an MLE.
Their derivation is based on analysis of an appropriate expected discrepancy (c.f.~\cite{linhart1986model}) for a Gaussian 
regression model.  In this section, we also follow this general approach, catering to our model.   

 Our model-based information criterion is obtained by appropriately penalizing 
the weighted log-likelihood of the multinomial model as specified in \eqref{eqn:base-form-of-Delta(K)}.
Specifically, we consider, for each $s$ and $\gamma > 0$, 
\begin{align}
\texttt{penalty}(K;s,\gamma) := \gamma \sum_{k=1}^K \frac{\widehat Z_k^{(K)}-1}{\left(\widehat N_k^{(K)}\right)^{s}}, \label{eqn:penterm-aicc}
\end{align}
where $\widehat \kappa^{(K)}$ is the estimate of  $\kappa$ assuming that $K^{*} = K$, 
$\widehat H^{(K)}$ is the matrix such that $\widehat H^{(K)}_{kt} := \bm 1\{\widehat \kappa^{(K)}(t) = k \}$,
$\widehat N_{k}^{(K)} := \sum_{t=1}^T N_t \widehat H_{k t}^{(K)}$,
$\widehat Z_{k}^{(K)} := \sum_{i=1}^{d} \bm 1\{\widehat Q^{(K)}_{ik} > 0\}$.  
In words, $\widehat N_{k}^{(K)}$ is the number of ``successes'' from the $k$th cluster specified by $\widehat \kappa^{(K)}$.
Also, $\widehat Z_{k}^{(K)}$ counts the number of non-zero entries from the $k$th cluster's success probability vector $\widehat Q_{k}^{(K)}$, 
as specified by $\mathcal {\widehat Q}^{(K)}$.

We detail our motivation for \eqref{eqn:penterm-aicc} by way of Theorem \ref{thm:result/main} and \ref{thm:main:ModelSelectThm}.
Intuitively, as $K$ increases, the term in \eqref{eqn:penterm-aicc} is expected to increase for a larger value of $K > K^{*}$ 
especially when $\widehat W_{i,c}^{(K)} >0$ and 
$\widehat W_{i,c^\prime}^{(K)} > 0$ for $c\neq c^\prime$ for many values of $i$.  In other words, when some columns of $\widehat W$ are ``overly'' similar to each other, the penalty term becomes more prominent (c.f.~Table \ref{tab:swimmer-result}).   

In Section \ref{sec:AICBICtoPen}, we reduce \eqref{eqn:penterm-aicc} to $\Delta_{\AIC}(K)$ and 
$\Delta_{\BIC}(K)$ in \eqref{eqn:DeltaAIC-classic}
and \eqref{eqn:DeltaBIC-classic} respectively, under some simplifying assumptions.     However, 
for clustering a sequence of sparse multinomial data, the penalty terms 
in \eqref{eqn:DeltaAIC-classic} and \eqref{eqn:DeltaBIC-classic} that are appropriate for classical normal regression problems, can over-penalize, especially  
when  the probability vectors $\mathcal Q$ contain many zeros (c.f.~Figure \ref{fig:c-elegan}).

\section{Theoretical results}
The main theoretical results of this paper are twofold.    First, we motivate a particular choice of 
the form of the penalty term in \eqref{eqn:penterm-aicc} through an asymptotic analysis of  $\Delta(K)$,
under simplifying assumptions that 
$\widehat \theta(K^{*}) = \theta^{*} \in \Theta(K^{*})$,  
$\widehat \theta(K^{*}-1) = \theta^{*,m} \in \Theta(K^{*}-1)$ and 
$\widehat \theta(K^{*}+1) = \theta^{*,s} \in \Theta(K^{*}+1)$.
Following \cite{1502.02069}, we use  the superscripted $m$ as  a mnemonic for ``merging'',
and use the superscripted $s$ as  a mnemonic for ``splitting''.
Second, we motivate the reduced rank projection approach 
for initializing the numerical search of the maximum $L_{q}$ likelihood estimate $\widehat \theta(K) \in \Theta(K)$.

\subsection{Asymptotic derivation of the penalty term}\label{sec:AsymDerivPenTerm}
In this section, we motivate a specific choice for the penalty term, $\texttt{penalty}(K)$, by computing the asymptotic form of
the expected weighted discrepancy of $\Delta(K^{*})$ while taking  the value of $\min_{t=1}^{T}\{N_{t}\}$ to $\infty$ along  
some sequence of index $\ell$.    Let
$\varphi(P) := -\sum_{t=1}^{T} \sum_{i=1}^{d} \frac{1}{N_t \overline n_{\kappa^*(t)}^2} \mathbf E\left[X_{i,t}\right] \log(P_{i,t}),$ 
where $P$ is associated with some $\theta \in \Theta$ through \eqref{eqn:NMFdecomposition000} 
and \eqref{eqn:NMFdecomposition111},
the expectation is taken with respect to $\theta^{*} \in \Theta(K^{*})$, whence 
$P_{it}^{*} = \frac{1}{N_t} \mathbf E\left[X_{i,t}\right]$, and  $\overline n_{k} = \sum_{t=1}^{T} \bm 1 \{ \kappa^{*}(t) = k \}$.
\begin{my.thm}\label{thm:result/main}
Suppose that
\begin{enumerate}
\item for each $\ell$, we have $N_{1}^{(\ell)}, \ldots, N_{T}^{(\ell)}$ such that 
for each $k=1,\ldots, K^{*}$, the number $N_t^{(\ell)}$ of trials is the same for all $t\in k(\kappa^{*})$,
\item for each $k=1,\ldots, K^*$,  there exists $\overline \lambda_{k} \in (0,\infty)$ such that 
for each $t \in k(\kappa^{*})$  
$$
\overline \lambda_{k} =\lim_{\ell}N_{t}^{(\ell)}/\ell.
$$
\end{enumerate}
Then,  
\begin{align*}
\lim_{\ell\rightarrow\infty} \ell \left( 
\mathbf E[\varphi(\widehat P) ] - \varphi(P^*) \right) = 
\frac{1}{2} \sum_{k=1}^{K^{*}} \frac{\overline Z_{k}^{*} -1}{\overline n_{k} \overline \lambda_{k}},
\end{align*}
where $\widehat P = X^{(\ell)} \diag(\bm 1^{\top} X^{(\ell)})^{-1}$ and
\begin{align}
\overline Z_k^{*} := \overline Z_k^{*}(Q^*) := \sum_{i=1}^{d} \bm 1\{W^*_{i,k} > 0\}.\label{eqn:Z term for counting non-zeros}
\end{align}
\end{my.thm}

Theorem \ref{thm:result/main} suggests $(1,1/2)$ for the value of the pair $(s,\gamma)$ in \eqref{eqn:penterm-aicc}. 
More importantly, we note that the non-zero entries do not contribute to the value of $\overline Z_k^*$ in \eqref{eqn:Z term for counting non-zeros}.

For the rest of this section, we further study, through Theorem \ref{thm:main:ModelSelectThm},
the question of for what values of $s$ and $\gamma$, we can expect to see that $\widehat K$ chosen 
according to \eqref{eqn:base-form-of-Delta(K)} with the penalty term specified by \eqref{eqn:penterm-aicc},
estimate the true value $K^{*}$ with high probability.   

Specifically, denoting by $N= \sum_{t} N_t$, in Theorem \ref{thm:main:ModelSelectThm}, we study the case in which
choosing $(s,\gamma)=(1/2,\log(N))$ will lead to a model selection criterion that tends
\begin{description}
\item[(i)] not to under-estimate the value of $K^{*}$ when the alternative model is one that the $(K^{*}-1)$st  and 
the $K^{*}$th clusters are ``merged'' into one,
\item[(ii)] not to over-estimate the value of $K^{*}$,
when the alternative model is one that for some $i$  and $j$,  $\kappa^{*,s}(i) = K^{*}$ and  $\kappa^{*,s}(j) = K^{*}+1$, i.e., 
the $K^{*}$th cluster is ``split'' into two. 
\end{description}

\begin{my.thm}\label{thm:main:ModelSelectThm}
Let $\theta^{*,m} = ( \kappa^{*,m}, \mathcal Q^{*,m}) \in \Theta(K^{*}-1)$ and 
$\theta^{*,s} = ( \kappa^{*,s},  \mathcal  Q^{*,s}) \in \Theta(K^{*}+1) $, where we let  
\begin{enumerate}
\item $ \kappa^{*,m}:\{1,\ldots, n\} \rightarrow \{1,\ldots,K^*-1\}$ be such that 
$ \kappa^{*,m}(i) = \kappa^{*}(i)$ for all $i$  that $\kappa^{*}(i) = k \le K^*-2$ and
$ \kappa^{*,m}(i) = K^*-1$ for all $i$ that $\kappa^{*}(i) = K^*-1$ or $K^*$,
\item $\mathcal Q^{*,m} = \{ Q^{*,m}_{k}\}_{k=1}^{K^*-1}$ 
be such that $ Q^{*,m}_{k}  = Q^{*}_{k}$ for $k=1,\ldots, K^*-2$ and  $ Q^{*,m}_{K^*-1} \in \mathbb R_{+}^{d}$ 
is a probability vector,
\item $ \kappa^{*,s}:\{1,\ldots, n\} \rightarrow \{1,\ldots,K^{*}+1\}$ be such that 
$ \kappa^{*,s}(i) = \kappa^{*}(i)$ for all $i$  that $\kappa^{*}(i) = k \le K^*-1$ and
$ \kappa^{*,s}(i) \in  \{K^{*},K^{*}+1\}$ for all $i$  that $\kappa^{*}(i) = K^{*}$, with $\kappa^{*,s}$ being surjective,
\item  $\mathcal Q^{*,s} = \{ Q^{*,s}_{k}\}_{k=1}^{K^{*}+1}$ be such that $ Q^{*,s}_{k}  = Q^{*}_{k}$ for $k=1,\ldots, K^*$ and then let $ Q^{*,s}_{K^{*}+1} = Q^{*}_{K^{*}}$.
\end{enumerate}
Suppose that $\lim_{N\rightarrow\infty}N_{t}/N > 0$ for each $t=1,2,\ldots, T$, and that
\begin{align}
& D_{KL}(\theta^{*}||\theta^{*,m}) \ge \min_{\theta \in \Theta(K^{*}-1)} D_{KL}(\theta^{*}||\theta) > 0, \\
&  D_{KL}(\theta^{*}||\theta^{*,s}) = \min_{\theta \in \Theta(K^{*}+1)} D_{KL}(\theta^{*}||\theta).
\end{align}
Then, 
\begin{align}
&\lim_{N\rightarrow\infty}\mathbf P[ \Delta^{*,m}(K^{*}-1) > \Delta^{*}(K^{*})] = 1, \label{eqn:AIC-lowerbound}\\
&\lim_{N\rightarrow\infty}\mathbf P[ \Delta^{*,s}(K^{*}+1) > \Delta^{*}(K^{*})] = 1,
\label{eqn:AIC-upperbound}
\end{align}
where
\begin{align*}
&\Delta^{*}(K^{*}) := \sum_{t=1}^{T} D_{t}(X_{t},  Q^{*}_{ \kappa^{*}(t)})  + \texttt{penalty}^{*}(K^{*}),\\
&\Delta^{*,m}(K^{*}-1) :=   \sum_{t=1}^{T} D_{t}(X_{t},  Q^{*,m}_{ \kappa^{*,m}(t)})  + \texttt{penalty}^{*,m}(K^{*}-1), \\
&\Delta^{*,s}(K^{*}+1) :=  \sum_{t=1}^{T} D_{t}(X_{t},  Q^{*,s}_{  \kappa^{*,s}(t)})  + \texttt{penalty}^{*,s}(K^{*}+1),
\end{align*}
with
\begin{align*}
&\texttt{penalty}^{*}(K^{*}) := \log(N) \sum_{k=1}^{K^{*}} \frac{\overline Z_{k}(Q^{*})-1}{(\sum_{t \in k(\kappa^*)} N_t)^{1/2}}, \\ 
&\texttt{penalty}^{*,m}(K^{*}-1) := \log(N) \sum_{k=1}^{K^{*}-1} \frac{\overline Z_{k}(Q^{*,m})-1}{(\sum_{t \in k(\kappa^{*,m})} N_t)^{1/2}},\\
&\texttt{penalty}^{*,s}(K^{*}+1) = \log(N) \sum_{k=1}^{K^{*}+1} \frac{\overline Z_{k}(Q^{*,s})-1}{(\sum_{t \in k(\kappa^{*,s})} N_t)^{1/2}}.
\end{align*}
\end{my.thm}

In other words, in Theorem \ref{thm:main:ModelSelectThm}, assuming that $\widehat \theta(K^{*}-1)$ is such that its $\widehat{\mathcal Q}$ takes a form of 
$\{Q^{*,m}_{1},\ldots, Q^{*,m}_{K^{*}-1}\}$ and its $\widehat \kappa$  takes a form of $\kappa^{*,m}$,  it follows that as $N\rightarrow\infty$, with high probability,
 $\Delta^{*,m}(K^{*}-1) > \Delta^{*}(K^{*})$, suggesting $\widehat K \ge K^{*}$.
Similarly, in Theorem \ref{thm:main:ModelSelectThm}, assuming that $\widehat \theta(K^{*}+1)$ is such that its $\widehat{\mathcal Q}$ takes a form of 
$\{Q^{*,s}_{1},\ldots, Q^{*,s}_{K^{*}+1}\}$ and its $\widehat \kappa$ takes a form of $\kappa^{*,s}$, and that
$\widehat \theta(K) = \theta^{*}$, it follows that as $N\rightarrow\infty$, with high probability,
 $\Delta^{*,s}(K^{*}+1) > \Delta^{*}(K^{*})$, suggesting $\widehat K  \le K^{*}$.
Also,   
 \eqref{eqn:base-form-of-Delta(K)} with the penalty term specified by \eqref{eqn:penterm-aicc} with $(s,\gamma) = (1/2,\log(N))$ yields the specific form 
of $\Delta^*(K^*)$, $\Delta^*(K^*-1)$ and $\Delta^*(K^*+1)$ in Theorem \ref{thm:main:ModelSelectThm}.

While proven under simplifying assumptions, 
through Theorem \ref{thm:result/main} and \ref{thm:main:ModelSelectThm},
we propose to choose  the value of $(s,\gamma)$
to be $(1/2,\log(N))$  for consistence estimation of $K^*$.

On the other hand, as discussed in Section \ref{sec:AICBICtoPen}, under some simplifying assumption,  $(s,\gamma) = (1,1)$ can be associated with the conventional AIC, and similarly, $(s,\gamma) = (1/2,\log(N)/\sqrt{N/T})$ can be associated with the conventional BIC.   For $(s,\gamma)=(1,1)$ and $(1/2,\log(N)/\sqrt{N/T})$, following the proof of Theorem \ref{thm:main:ModelSelectThm},  one can show results similar to
\eqref{eqn:AIC-lowerbound} while the probability in \eqref{eqn:AIC-upperbound} is positive but can be strictly less than $1$.  
For our numerical experiments in Section \ref{sec:NumericalResult},  to be comparable to the conventional AIC, 
we take $(s,\gamma) = (1,1)$ and  we give a preference to a smaller value for $\widehat K$ if a near-tie occurs.

\subsection{Reduced rank projection as a smoothing routine}
The motivation behind using a  reduced rank projection step is to remove random variation. 
As discussed in \cite{kmo10noise}, when there is no missing entries in $X$, $\OptSpace$ algorithm is equivalent to performing reduced-rank projection (or equivalently, singular value thresholding at a fixed rank).
Specifically, it can be seen from \cite[Theorem 4.4]{kmo10noise}, that provided that a random matrix $M$ is assumed to be bounded appropriately
and that its entries $\{M_{ij}\}$ are independent random variables, using $\OptSpace$ yields a consistent estimate of $\mathbf E[M]$ under various conditions. 

To give a more precise statement of our contribution on the topic,  
we begin by introducing some notation. 
Given a constant $C>0$, for each $i$ and $t$, let
\begin{align*}
Y_{i,t} :=  X_{i,t} \wedge C := \min\{X_{i,t}, C\}.
\end{align*}
Then, we let $\widehat Y$ be the result of a single iteration 
of the singular value threholding of $Y$ using the (true) value of $K^{*}$ of 
the matrix $\mathbf E[X]$.   
We will suppress, in our notation, the dependence of $X$, $Y$, 
and $\widehat Y$ on $C,n,T$ for simplicity.

Truncating each $X_{i}$ at $C$ yields an estimate that 
is biased due to truncation while its effect may diminish 
as the value of $C$ increases. We present 
an asymptotic result in which $C$ is allowed to grow as a 
function of $d$ and $T$ under several simplifying assumptions.

Our first simplifying assumption is that the mean matrix $\mathbf E[X]$ has 
a ``finite'' block structure, or equivalently, a ``finite'' checker-board type pattern.
Specifically, we assume that
$ \mathcal B_1,  \ldots , \mathcal B_B$ partition
the index set  $\mathcal B := \{(i,t) : i  = 1,\ldots, d, t=1,\ldots, T\}$,
where the value of $B$ is constant and does not depend on $C$, $d$ and $T$,
Next, we assume also that for each $b=1,\ldots, B$, there exists a pair $(\nu_{b},p_{b}) \in (0,\infty) \times (0,1)$ such that for each $(i,t) \in \mathcal B_b$,
$\mathbf E[X_{i,t}]= \nu_b$ and $\lim_{d\wedge T\rightarrow\infty} |\mathcal B_b|/|\mathcal B| = p_b$.
We assume that the values of $\{ (\nu_{b},p_{b}): b=1,\ldots, B\}$ are constant and do not depend on $C$, $d$ and $T$.

We suppress in our notation, the dependence of 
$\nu_b$, $\mathcal B_b$, $\mathcal B$ and $C$
on $d$ and $T$ for simplicity.  Also, $d \wedge T \rightarrow \infty$ 
means that the pair $(d,T)$ is indexed by $\ell=1,2,\ldots$, so
that $\lim_{\ell\rightarrow\infty} \min(d_\ell,T_\ell) =\infty$.  

\begin{my.thm}\label{thm:meantotalinteraction}
Suppose that $N_{1}, N_{2}, \ldots, N_{T}$ are independent Poisson random variables, and that the rank of $\mathbf E[X]$ is $K^{*}$.

If $\lim_{T\wedge d\rightarrow\infty} C_{d,T} =\infty$ and $C := C_{d,T}=o((d/T^{3})^{1/4}) \wedge o(\log(T)^{1/2}/T)$, 

then
\begin{align*}
\lim_{T \wedge d \rightarrow\infty }\MSE(\widehat Y;X) = 0,
\end{align*}
where $\MSE(\widehat Y;X) := \mathbf E\left[ \frac{1}{dT} \|\widehat Y - \mathbf E[X]\|_F^2\right]$.
\end{my.thm}

Note that generally, the rank of $\mathbf E[X] \le K^{*}$ and 
for some cases, it is also possible to have the rank of $\mathbf E[X] < K^{*}$.
In Theorem \ref{thm:meantotalinteraction}, to simplify our analysis,
we have assumed that the rank of $\mathbf E[X]$ is $K^{*}$.
Next, to consider Theorem \ref{thm:meantotalinteraction} with respect to 
\cite[Theorem 4.4]{kmo10noise}, we note that
the result in \cite[Theorem 4.4]{kmo10noise} applies 
when the errors are independent while the entries of $\widehat P$ are correlated.  
Specifically, as a corollary  to Theorem \ref{thm:meantotalinteraction}, we also have that, under the hypothesis of 
Theorem \ref{thm:meantotalinteraction}, 
\begin{align}
\lim_{T \wedge d \rightarrow\infty } \frac{1}{dT} \mathbf E[\| \widetilde P - P^{*} \|_{F}^{2} ]    = 0
\end{align}
where $\widetilde P = Y \diag(\bm 1^{\top} Y)^{-1}$, since given the value of $\bm N = (N_{1}, N_{2},\ldots, N_{T})$, 
\begin{align*}
& \mathbf E[\| \widetilde P - P^{*} \|_{F}^{2} \left|\bm N \right.]    
\le  \frac{1}{\min_{t=1}^{T} N_{t}^{2}}  \mathbf E[\| \widehat X -   \mathbf E[X] \|_{F}^{2} \left| \bm N \right.],
\end{align*}
where $\bm N = (N_{t})_{t=1}^{T}$.
In this manner, in addition to giving a motivation to reduced rank projection as a smoothing routine, 
Theorem  \ref{thm:meantotalinteraction} can be of interest on its own.

\section{Numerical results}\label{sec:NumericalResult}

\subsection{Simulation experiments}
\begin{table}   
\centering
\caption{Comparison of $\Delta(K)$ and $\Delta_{\AIC}(K)$ in terms of the values of $\Delta(K)$ and $\Delta_{\AIC}(K)$, for a single-instance of a $50\times 2$ data matrix generated using a two-cluster parameter, i.e., $K^{*}=2$.}\label{tab:comparativeMLqE}
\begin{tabular}{ccccccc}
  \hline
$K$ & $D(\widehat{\mathcal Q},  \widehat \kappa)$ & \texttt{penalty} & $\Delta(K)$ & \texttt{penalty} & $\Delta_{\AIC}(K)$ \\ 
  \hline
1 & 22.18 & 0.01 & 22.18 & 0.02 & {\bf 22.20} \\ 
  2 & 22.14 & 0.02 & {\bf 22.16} & 0.08 & 22.22 \\ 
   \hline
\end{tabular}
\end{table}

\subsubsection{Simple experiment}
In this section, through a simple numerical experiment, 
we reiterate our last observation made in Section \ref{sec:ModelSelection}.
Consider a sequence $(X_{1}, X_{2})$ of multinomial random vectors taking 
values in $\{0,1,2,\ldots, \}^{50}$, where their (common) number $N_{t}$ of multinomial trials  is $200$.  
Specifically, the first success probability vector is proportional to the vector $(1,\ldots, 1, 10, 10,10,10,0,\ldots, 0) \in \mathbb R^{50}$ and the second probability vector is proportional to the vector $(0,\ldots, 0, 10, 10,10,10,1,\ldots, 1) \in \mathbb R^{50}$, where for both cases, the number of entries with its value being $0$ is $23$ and the number of entries with its value being $1$ is $23$. 
In other words, the value of $K^{*}$ is $2$, whence in this case, our model selection procedure seeks
to reach the minimum value of $\Delta(K)$ with $K=2$.   
As shown in Table \ref{tab:comparativeMLqE}, the value of $\Delta(K)$ is minimized at $K=2$ while 
the value of $\Delta_{\AIC}(K)$ is minimized at $K=1$.   

More generally, in Table \ref{tab:comparativeNewAICvsOldAIC}, we allow the value of $d$ to grow, while keeping the first success 
probability vector to be a scalar multiple of the vector $(1,\ldots, 1, 10, 10, \cdots,10,0,\ldots, 0) \in \mathbb R^{d}$
and  keeping the second success 
probability vector to be a scalar multiple of $(0,\ldots, 0, 10, 10,\cdots,10,1,\ldots, 1) \in \mathbb R^{d}$, where for both cases, the number of entries with its value being $10$ is fixed at $10$ and the number of entries with its entries being $0$ and $1$ are  the same or differ only by $1$.  A general pattern Table \ref{tab:comparativeNewAICvsOldAIC} is that  for all values of $d$, in comparison to $\Delta$, the conventional AIC, i.e. $\Delta_{\AIC}$, performs poorly, and we attribute this to the fact that $\Delta_{\AIC}$ over-penalizes in comparison to $\Delta(K)$.  

 \begin{table}
 \centering
 \caption{Comparison of $\Delta$ and $\Delta_{\AIC}$ in terms of the number of times that $\widehat K = 2$ out of $100$ Monte Carlo repetitions.  For each $d = 20, 25,\ldots, 100$, each Monte Carlo replicates of a $d\times 2$ data matrix is  generated using a two-cluster parameter.}\label{tab:comparativeNewAICvsOldAIC}
 \begin{tabular}{cccc}
   \hline
 $d$ & $\Delta$ & $\Delta_{\AIC}$ \\
   \hline
 20 & 11 & 0 \\
   25 & 61 & 1 \\
   30 & 86 & 6 \\
   35 & 100 & 16 \\
   40 & 99 & 25 \\
   45 & 100 & 52 \\
   50 & 100 & 56 \\
   55 & 100 & 60 \\
   60 & 100 & 70 \\
   65 & 100 & 70 \\
   70 & 100 & 78 \\
   75 & 100 & 72 \\
   80 & 100 & 83 \\
   85 & 100 & 76 \\
   90 & 100 & 81 \\
   95 & 100 & 80 \\
   100 & 100 & 82 \\
    \hline
 \end{tabular}
 \end{table}

\subsubsection{Comparison to rank determination strategies}
We now present numerical results for comparing our approach 
to two conventional rank determination methods.  Specifically, 
we denote our first baseline algorithm with (pamk o dist) and the second with (mclust o pca), where o denotes composition.  
These competing algorithms are often used in practice for choosing the rank of a (random) matrix.   In comparison, we denote 
our model selection procedure by (aic o nmf).    
For (pamk o dist), one first computes  
the distance/dissimilarity matrix using pair-wise Euclidean/Frobenius distances between the columns of $X$,
and perform  \emph{partition around medoids} for clustering (c.f.~\cite{dudaHart1973}) together with ``Silhouette'' criterion (c.f.~\cite{rousseeuw1987silhouettes}) for deciding the number of clusters. 
For (mclust o pca), one first uses  an ``elbow-finding'' algorithm to estimate the rank $K^{*}$ of the data matrix $X$ (c.f.~\cite{ZhuGhodsi2006}), say, by $r$, and then, use a clustering algorithm (c.f.~\cite{mclust2002}) to cluster columns of $X$ into $r$ clusters.

\begin{figure}
\centering
\includegraphics[width=0.45\textwidth]{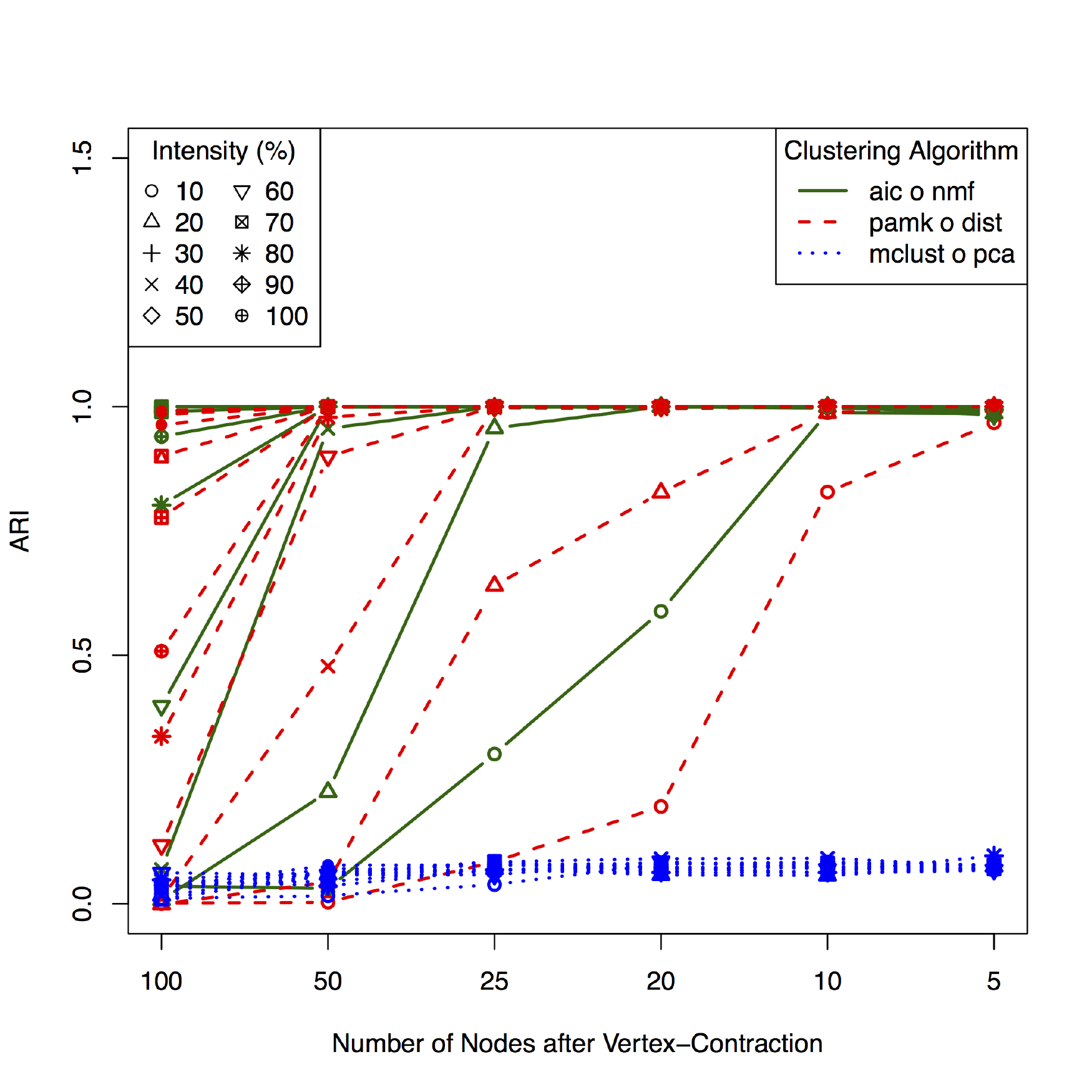}
\caption{Comparison of three  approaches through ARI for  the model selection performance.  
In all cases, our procedure either outperforms or nearly on par with the two baseline algorithms. 
}\label{fig:ARIasNodeSparsity}
\end{figure}

The result of our  experiment is summarized in Figure \ref{fig:ARIasNodeSparsity}, which illustrates that  
in all cases, (aic o nmf) either outperforms or nearly on par with the two baseline algorithms. 

To explain our result, we now specify the  set-up for our Monte Carlo experiment.
Our experiment is motivated by the real data experiment studied in Section 
\ref{sec:c-elegan}, where a problem of comparing two graphs representing electrical and chemical neuron pathways of C.~elegan is studied. 

Specifically, we consider random graphs on $n$ vertices such that  each $\mathbf E[G(t)]$
has a block-structured pattern, i.e., a checker-board like pattern (c.f.~Figure \ref{fig:c-elegan}).  
For each $t=1,\ldots, T$, 
we take $G(t)$ to be a (weighted) graph on $n$ vertices, where each $G_{ij}(t)$ is a Poisson random variable. 
To parameterize the $5\times 5$ block structures, we set the number of vertices 
$n = 5 \times m$, where $m=20$.

The value of $m$ equals the number of rows in each block. 
Then, given a value for the intensity parameter $\rho \in [0,1]$, 
for each $i$ and $j=1,\ldots, n$, 
we let $\mathbf E[G_{ij}(t)] = 100 \rho B_{uv}^{(t)}$, where  $u$ and $v \in \{1,2,\ldots, 5\}$ are such that  
$20(u-1)+1 \le i \le 20 u$ and $20(v-1)+1 \le j \le 20 v$. 
We take
\begin{align*}
& B^{(1)} :=
\begin{pmatrix}
0.1  & 0.045 & 0.015 & 0.19 & 0.001\\
0.045 & 0.05 & 0.035 & 0.14 & 0.03\\
0.015 & 0.035 & 0.08 & 0.105 & 0.04\\
0.19 & 0.14 & 0.105 & 0.29 & 0.13\\
0.001    & 0.03 & 0.04 & 0.13 & 0.09
\end{pmatrix}, \\
&B^{(2)} :=
\begin{pmatrix}
0.19 & 0.14 & 0.29  & 0.105 & 0.13 \\
0.001& 0.03 & 0.13  & 0.04  & 0.09 \\
0.015& 0.035& 0.105 & 0.08  & 0.04 \\
0.045& 0.05 & 0.14  & 0.035 & 0.03 \\
0.1  & 0.045& 0.19  & 0.015 & 0.001
\end{pmatrix}.
\end{align*}

In Figure \ref{fig:ARIasNodeSparsity}, the horizontal axis specifies the number $c$ of nodes 
after aggregation.  
For the level of aggregation (or equivalently, vertex-contraction), 
if the number of nodes after vertex-contraction is $5$ (i.e. the far right side of Figure \ref{fig:ARIasNodeSparsity}), the original graph is 
reduced to a graph with $5$ vertices.  Aggregation of edge weights is only done 
within the same block.  
Then, we take $X$ to be the $c^{2} \times 2$ non-negative matrix such that its $t$th column is the vectorized version of the aggregation of $G(t)$.  
Our problem is then to estimate the number of clusters 
using data $X$, where the correct value for $\widehat K$ is $K^{*}=2$. 
In this particular case, $\mathbf E[X]$ is a rank-$2$ matrix, and as such, our problem can also 
be thought to be a problem of estimating the rank of $\mathbf E[X]$ after observing $X$, whence 
(pamk o dist) and (mclust o pca) are applicable.

In summary, there are two parameters that we have varied, specifically, the level of intensity and the level of aggregation. 
The level of intensity is changed by the value of $\rho \in (0,1)$, which is distinguished in Figure \ref{fig:ARIasNodeSparsity} by the shape of points.   
Note that a bigger value for $\rho$ means more chance for each entry of $X$ taking a large integer value.  

Then, as the performance index, we use 
the adjusted Rand index (ARI) values (c.f.~\cite{WMRand1971}).  In general, ARI takes a value in $[-1,1]$. 
The cases in which the value of ARI is close to $1$ is ideal, indicating that clustering is consistent with the truth, and 
the cases in which the value of ARI is less than $0$ are the cases in which its performance is worse than randomly assigned clusters.   Then, to compare three algorithms,  we compare the values of ARI 
given each fixed value of $(\rho, c) \in [0,1] \times \{100,50, \ldots, 5\}$.

\subsubsection{Comparison to a non-parametric two-sample test procedure for random graphs}
In this section, we consider a sequence of undirected loop-less (unweighted) 
random graphs $G_{1}, \ldots, G_{T}$ such that 
$\mathbf E[G_{t}]$ is an element of a set of $K^{*}$ distinct $n\times n$ matrices whose entries are probabilities.
Then, we consider a problem of clustering $T$ graphs into finite number of groups from the data $G_{1},\ldots, G_{T}$.
This is an abstraction of a problem in neuroscience, where each $G_{t}$ can represents a copy of neuron-to-neuron
interaction pattern, where each $G_{t}$ portraits a different mode of connectivity between neurons.  For instance, in Section 
\ref{sec:c-elegan}, the modes are the chemical and the electrical pathways. 

Since each $G(t)$ is undirected and loop-less, its adjacency matrix can be embedded as a vector 
$X_{t}$ in an element in $\{0,1\}^{n(n-1)/2}$.  
For our simulation study, we take $T=6$, and take $P_{1} = P_{2} = P_{3} = Q_{1}$  and $P_{4} = P_{5} = P_{6} = Q_{2}$.  Then, each $X_t$ takes values in 
$\{0,1\}^{d}$, where $d = {100\choose 2}=4950$ for $n=100$.   Hence, each $Q_t$ is a vector in $(0,1)^{d}$.    For our simulation 
study, we let $n$ to take values in $\{40, 60, 80, 100, 120\}$. 

For each $t=1,\ldots,T=6$, the matrix $M_{t} = \mathbf E[G_{t}]$ is 
an $n\times n$  block-patterned symmetric matrix such that each $M_{t,ij} \in 
(0,1)$ and each $G_{t,ij}$ is a Bernoulli random variable with success 
probability $M_{t,ij}$.   Put differently, we simulate each $G_{t}$ according to a (degenerate) stochastic block model, which generalizes the celebrated Erdos-Reyni random graph. The stochastic block model owes its popularity for being a model useful in practice while still being analytic, and we direct the reader's attention to \cite{1502.02069} and \cite{pavlovic2014stochastic} for a more detailed treatment.

For our simulation study with $n=100$, we take 
\begin{align*}
M_{1} = 
\begin{bmatrix}[c|c|c|c]
B_{11} & B_{12} & B_{13} & B_{14} \\ \cline{1-4}
B_{21} & B_{22} & B_{23} & B_{24} \\ \cline{1-4}
B_{31} & B_{32} & B_{33} & B_{34} \\ \cline{1-4}
B_{41} & B_{42} & B_{43} & B_{44}
\end{bmatrix},
\end{align*}
where each $B_{uv}$ is a $25 \times 25$ matrix such that its entries assume
the same value $b_{uv}$.  Moreover, we set
\begin{align*}
[b_{uv}] = 
\begin{bmatrix}
0.75 & 0.25 & 0.25 & 0.25 \\  
0.25 & 0.75 & 0.25 & 0.25 \\   
0.25 & 0.25 & 0.75 & 0.25 \\  
0.25 & 0.25 & 0.25 & 0.75 
\end{bmatrix}.
\end{align*} 
Next, we take 
\begin{align*}
M_{2} = 
\begin{bmatrix}[c|c|c]
B_{11}^{\prime} & B_{12}^{\prime} & B_{13}^{\prime} \\  \cline{1-3}
B_{21}^{\prime} & B_{22}^{\prime} & B_{23}^{\prime} \\  \cline{1-3}
B_{31}^{\prime} & B_{32}^{\prime} & B_{33}^{\prime}
\end{bmatrix},
\end{align*}
where the entries of each block $B_{uv}^{\prime}$ assume
the same value $b_{uv}^{\prime}$, 
$B_{11}^{\prime}, B_{13}^{\prime}, B_{31}^{\prime}$ are $25 \times 25$ matrices, and  
$B_{12}^{\prime}$ and $(B_{23}^{\prime})^{\top}$ are $25 \times 50$ matrices.  
Moreover, we set
\begin{align*}
[b_{uv}^{\prime}] = 
\begin{bmatrix} 
0.6 & 0.4 & 0.4 \\    
0.4 & 0.6 & 0.4 \\  
0.4 & 0.4 & 0.6
\end{bmatrix}.
\end{align*}
Note that, for $M_{1}$,  the set of $100$ vertices is partitioned into $4$ groups, 
and for $M_{2}$, the set of $100$ vertices is partitioned into $3$ groups, where
the first and the last groups are composed of $25$ vertices, and the middle group 
is composed of $50$ vertices.  

The adjusted Rand index is used to compare the clustering result of our approach to 
the ground truth, i.e., $\widehat K$ versus $K^{*}$, and $\widehat \kappa$ versus $\kappa^{*}$.
For each $n$, 
$100$ Monte Carlo clustering experiments were performed, yielding $100$ adjusted Rand index values, which were averaged.
As the number $n$ of vertices takes values in $\{40,60,80,100,120\}$, the average of the values of adjusted Rand index
from $100$ Monte Carlo experiments, took values in $\{0.42, 0.6, 0.8, 0.9, 0.92\}$ respectively.

To put the aforementioned numeric result in a context, we compare our approach to a non-parametric two-sample test approach
of \cite{1409.2344} for comparing graphs.  To be self-contained, we briefly outline the steps of the two-sample test approach 
of \cite{1409.2344} for comparing graphs.  Specifically, first,
using the singular value decomposition of \emph{each} $G(t)$, $n$ vertices were embedded as $n$ points $(Y_{i}(t))_{i=1}^{n}$ 
in a Euclidean space with its dimension much less $n$, and then for each pair $(t, s)$ with $1 \le t < s=T=6$,  
the technique in \cite{1409.2344} was used to compute the $p$-value $p(t,s)$ for testing whether or not  
the (empirical) density for $(Y_{i}(t))_{i=1}^{n}$  and the (empirical) density for  $(Y_{i}(s))_{i=1}^{n}$ 
are identically distributed.  Then,  define $D$ to be the $6\times 6$ hollow symmetric matrix such that 
$D_{ts} = p(t,s)$ for $t < s$, and subsequently, a technique akin to the principal component analysis is applied to $D$ and then a $K$-means clustering algorithm
is used to cluster six ``graphs''.   For a more detailed description and analysis of the algorithm, we direct the reader to \cite{1409.2344}, where the testing procedure is shown to be consistent as $n\rightarrow\infty$.  
As before, for each $n$, 
$100$ Monte Carlo clustering experiments were performed, yielding $100$ adjusted Rand index values, which were averaged.
As the number $n$ of vertices takes values in $\{40,60,80,100,120\}$, the average of the values of adjusted Rand index
from $100$ Monte Carlo experiments, took values in $\{0.18,0.28,0.33,0.38,0.39\}$ respectively.

\begin{table}   
\centering
\caption{Averaged value of ARI from a Monte Carlo experiment comparing two graphs}\label{tab:twosample-ari}
\begin{tabular}{cccccc}
\qquad & 40 & 60 &80 &100 &120 \\ \hline
aic o nmf  & 0.42   & 0.6   &   0.8 & 0.9   & 0.92 \\
two sample & 0.18   & 0.28 & 0.33 & 0.38 &0.39 \\ \hline
\end{tabular}
\end{table}

As can be seen in Table \ref{tab:twosample-ari}, our approach outperforms the non-parametric two-sample approach for each $n=40,60,80,100,120$.   On the 
other hand, this is, to some extend, understandable because the non-parametric  two-sample approach uses embedding, and after embedding 
the algorithm ignores the information that the $i$th vertex in $G(t)$ is also the $i$th vertex in $G(s)$ for any $t < s$. 
In other words, an advantage of the non-parametric  two-sample algorithm is that it can apply to a problem even when the vertex 
correspondence between vertices of $G(t)$ and the vertices of $G(s)$ is unknown, 
but in our present situation, the advantage becomes 
a disadvantage.  

To this end, to make a more fair comparison, we modify our original problem slightly so that 
given $n$ vertices, we are allowed to assume the knowledge of the vertex correspondence across $T$ graphs only for some vertices.   Then, for our approach only, to rectify the issue of unknown correspondence, we apply the technique
known as the ``seeded'' graph matching algorithm of \cite{2013arXiv1310.1297L} to best extrapolate the unknown correspondence, before applying 
our approach.  Specifically, given the number $n$ of vertices, we step the number $m$ of the known 
vertices toward $n$ in an increment of $5$.   

\begin{figure}
\centering
\includegraphics[width=0.5\textwidth]{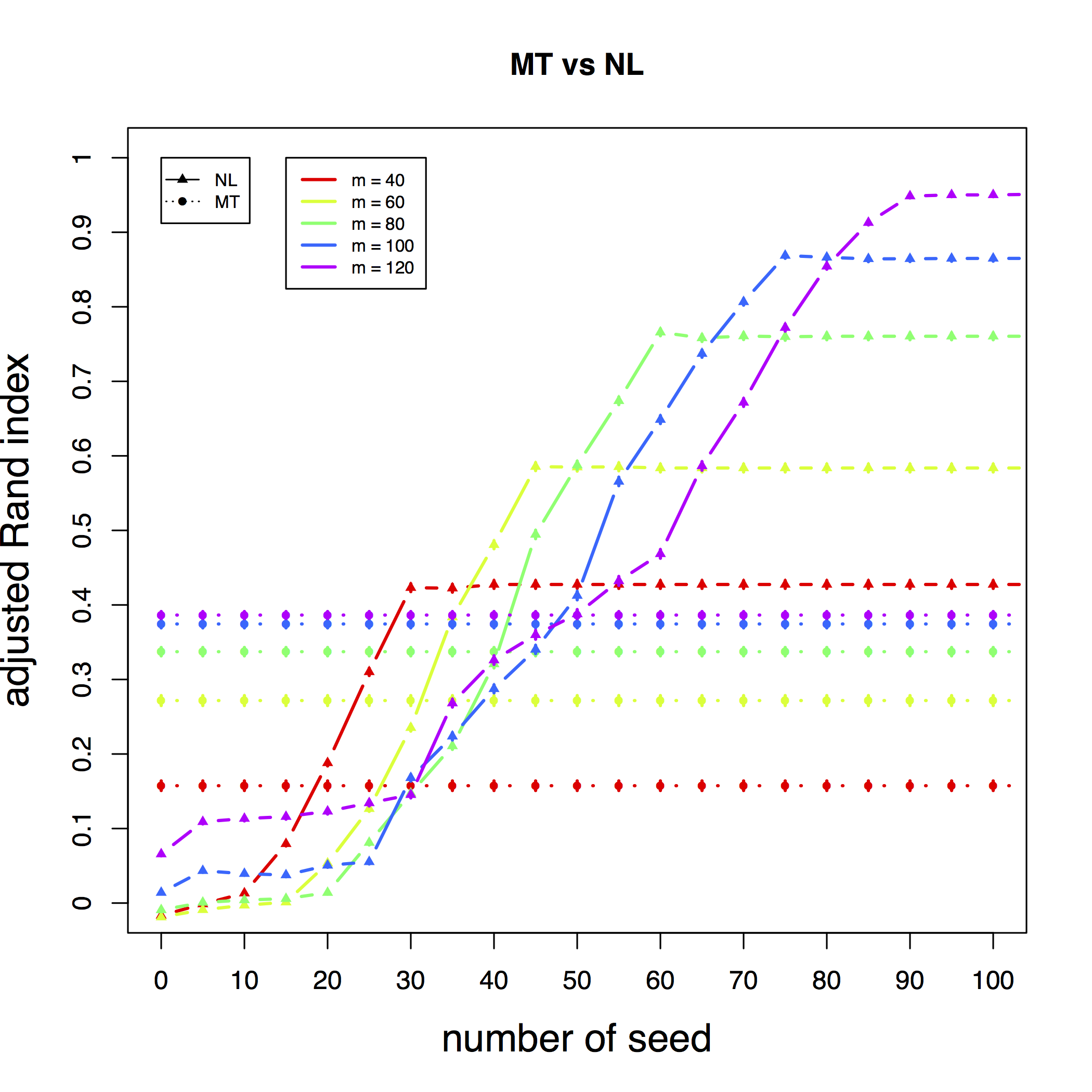}
\caption{Comparison of two clustering algorithms, specifically, the two-sample test procedure (MT) in \cite{1409.2344} and our clustering approach (NL)  which is (aic o nmf)   When the vertex correspondence is fully known, our approach outperforms, but when 
no vertex correspondence is known, the non-parametric  two sample procedure outperforms.  When only fraction of the true correspondence is known, the unknown correspondence is extrapolated from the known correspondence from the connectivity pattern, using the ``seeded'' graph matching algorithm from \cite{2013arXiv1310.1297L}.}\label{fig:MTvsNL}
\end{figure}

The result is illustrated in Figure \ref{fig:MTvsNL}, 
where for compactness, \emph{MT} abbreviates \emph{Multiple hypothesis Testing }for the non-parametric two-sample approach,
and \emph{NL} abbreviates \emph{Non-zero penalizing weighted Likelihood} for our model selection criterion, i.e., (aic o nmf).
In summary, 
for each $n$, when $m$ is small, the non-parametric two-sample test approach outperforms our approach, but when $m$ is moderate or large, 
our approach outperforms the non-parametric two-sample test approach.  The low values of the adjusted Rand index for our approach when $m/n \approx 0$ and for the non-paramaetric two-sample test approach when $m/n \approx 1$ are understandable.  We conjecture that 
the location at which the values of the adjusted Rand index for two approaches cross over is a property of the underlying 
``seeded graph matching'' algorithm of \cite{2013arXiv1310.1297L}, but a deep analysis of this phenomenon is beyond the scope of our present work.

\subsection{C.~elegan connectomics}\label{sec:c-elegan}
\begin{figure}
\centering
\includegraphics[width=0.425\textwidth]{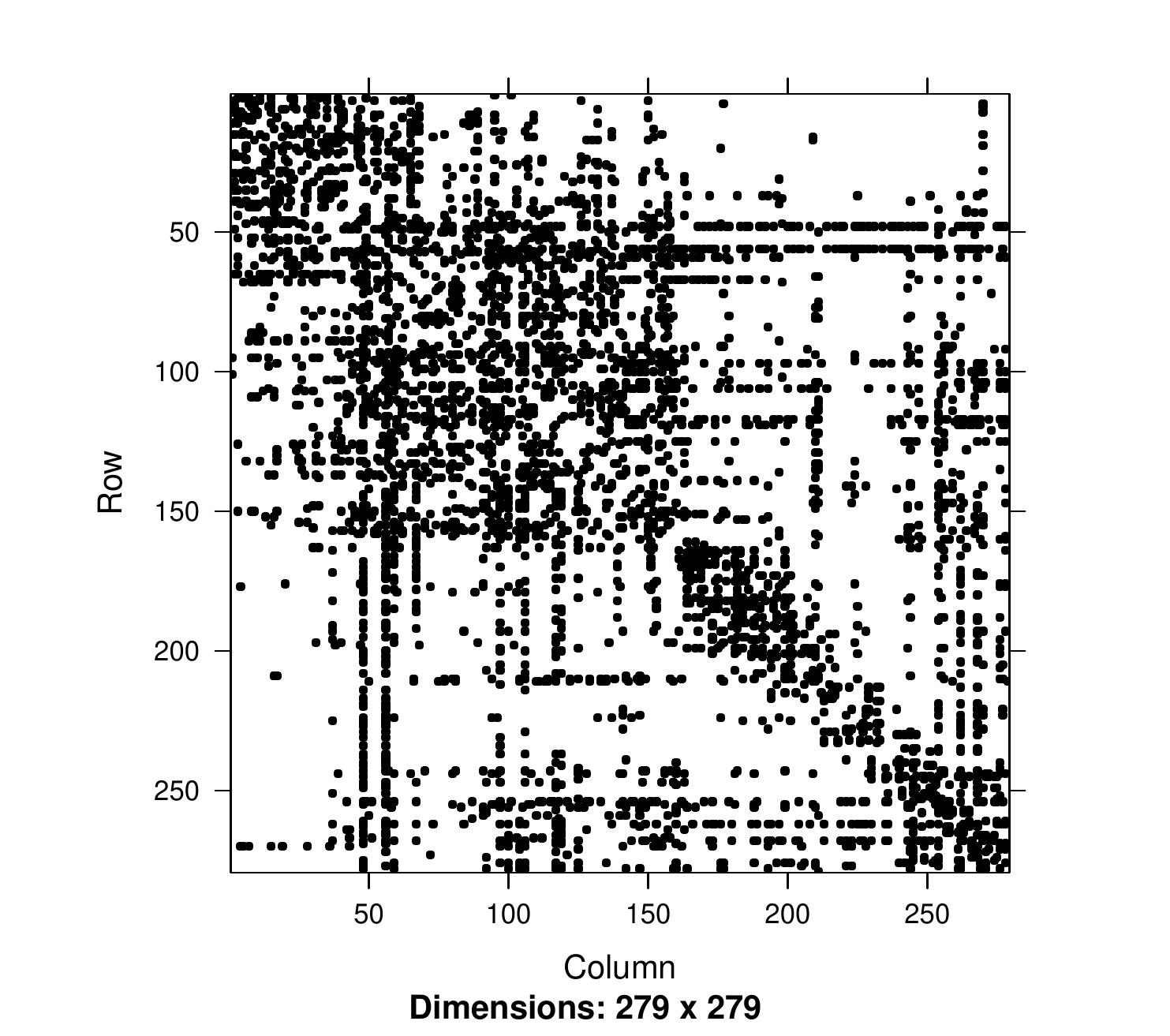}
\includegraphics[width=0.425\textwidth]{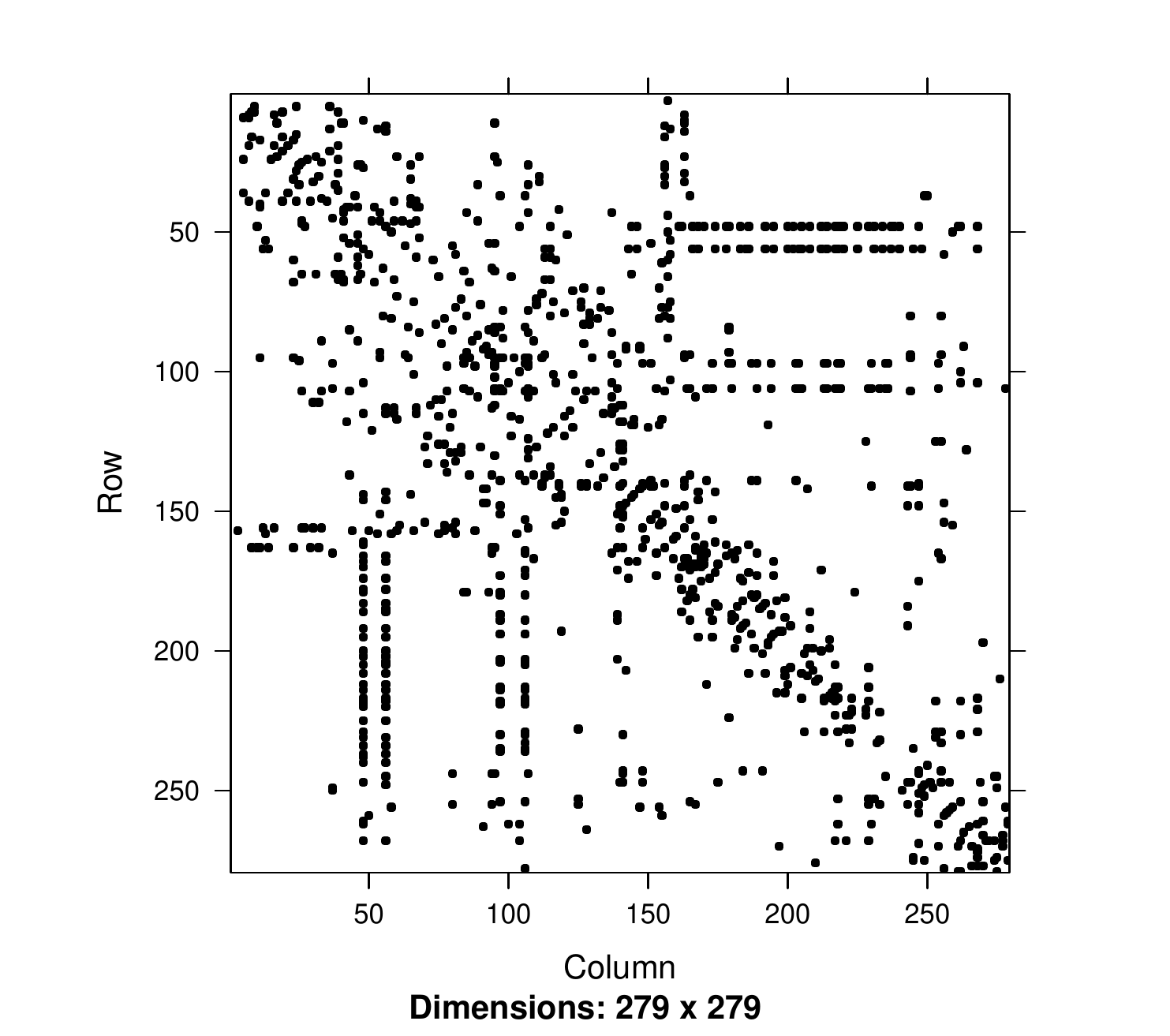}
\caption{Connectivity matrices for C.~elegan's chemical and electrical networks between neurons.  Visually, while there are  similarities between
the graphs representing two networks, it can be seen that there are also dissimilarities.  Our numerical 
experiment yields that $\widehat K = 2$, further corroborating that two networks are sufficiently different.} \label{fig:c-elegan}
\end{figure}

In \cite{jarrell2012connectome}, to study the decision-making process of the C.~elegan, 
chemical and electrical neuronal pathways of a C.~elegan worm's  $279$ neurons were observed, 
yielding a pair of graphs on the same (matching) vertex set. 
First, $279$ neurons are collapsed according to their types, yielding a pair of graphs on $3$ vertices.
This yields a $3\times 2$ matrix $X$, where each row corresponds to a pair of vertices (collapsed neurons), 
and the two columns correspond to two types of pathways.   
Our clustering approach yields that the value of  $\Delta(K)$ for $K=1$ and $K=2$ are $7.79$ and $7.69$,
suggesting $\widehat K = 2$.   
Next, to allow for a larger dimension while avoiding working with an excessively sparse matrix,
vertex contraction is performed so that for each of the first eight groups of thirty neurons, 
its thirty neurons are aggregated (collapsed)  to a single vertex, 
and then, the remaining thirty-nine vertices are aggregated to a single vertex.  
These groupings do not signify any special feature.   
This yields a pair of weighted graphs on $9$ vertices, whence the corresponding matrix $X$ is $36\times 2$ matrix $X$, because 
$36 = {9 \choose 2}$.
Performing our clustering procedure to the matrix yields that  the values of $\Delta(K)$ for $K=1$ 
and $K=2$ are $15.84$ and $15.61$, suggesting that there are two patterns. 
In words, our approach suggests that the chemical pathways and electric pathways 
of the C.~elegan worm were sufficiently different with respect to their connectivity patterns during the period of observation, corroborating 
the visual patterns observed in Figure \ref{fig:c-elegan}.

\subsection{Swimmer Dataset}\label{[exa:swimmer+visual.tfidf]}
The swimmer data set is a frequently-tested data set for bench-marking 
NMF algorithms (c.f.~\cite{Donoho03whendoes} and \cite{NickGillis2014}). 
In our present notation, each column of $220\times 256$ 
data matrix $X$ 
is a vectorization of a binary image, and each row corresponds to 
a particular pixel.  
Each image is a binary images ($20$-by-$11$ pixels) 
of a body with four limbs which can be positioned in four different 
positions.  As such, in the language of \cite{NickGillis2014}, it can be seen 
that the matrix $X$  is $16$-separable, or equivalently, the (minimal) inner dimension of $X$ 
is $16$.   However, as it so happens, 
the rank of $X$ is $13$.  In other words, there are $16$ basic 
patterns/motifs in $X$ that are repeated, and
the rank of $X$ being $13$ is a nuisance fact.
As displayed in Table \ref{tab:swimmer-result}, the value of $\Delta(K)$ is minimized at 
$K =16$, matching the inner dimension of $X$.

\begin{table}   
\centering
\caption{Values of $\Delta(K)$ for estimating the inner dimension for the Swimmer dataset}\label{tab:swimmer-result}
\begin{tabular}{cccc}
  \hline
$K$ & $D(\widehat{\mathcal Q},  \widehat \kappa)$ & \texttt{penalty} & $\Delta(K)$  \\ 
  \hline

 1 & 997.87 & 0.00 & 997.87 \\
   2 & 986.32 & 0.02 & 986.34 \\
   3 & 982.11 & 0.05 & 982.16 \\
   4 & 982.94 & 0.08 & 983.02 \\
   5 & 982.16 & 0.10 & 982.27 \\
   6 & 979.66 & 0.14 & 979.80 \\
   7 & 973.32 & 0.16 & 973.49 \\
   8 & 967.80 & 0.20 & 967.99 \\
   9 & 958.16 & 0.27 & 958.42 \\
   10 & 957.31 & 0.31 & 957.62 \\
   11 & 957.71 & 0.31 & 958.02 \\
   12 & 947.05 & 0.35 & 947.40 \\
   13 & 907.26 & 0.32 & 907.58 \\
   14 & 889.82 & 0.51 & 890.32 \\
   15 & 865.97 & 0.76 & 866.73 \\
{\bf   16} & {\bf 834.69} & {\bf 0.91} & {\bf 835.59} \\
   17 & 834.66 & 4.07 & 838.73 \\
   18 & 834.65 & 13.79 & 848.44 \\
   19 & 834.65 & 29.27 & 863.92 \\
   20 & 834.65 & 10.61 & 845.27 \\    \hline
  \end{tabular}
\end{table}

\section{Discussion}\label{sec:discussion}
Theorem \ref{thm:result/main}, \ref{thm:main:ModelSelectThm}, and 
Theorem \ref{thm:meantotalinteraction} are proven under simplifying assumptions.  
Specifically, a shortcoming of Theorem \ref{thm:meantotalinteraction} is that $\mathbf E[X]$ has 
a finite dimensional block structure, while this can be relaxed to 
other various settings in which the number of blocks can grow.  
Next, a shortcoming of Theorem \ref{thm:result/main}
and \ref{thm:main:ModelSelectThm} is that 
our analysis is done with $Q^{*}$, $Q^{*,m}$ and $Q^{*,s}$ rather than their $\MLqE$ counter-parts.  A remedy for this shortcoming is to use a concentration-inequality type argument to show that their counter-parts are concentrated at $Q^{*}$, $Q^{*,m}$ and $Q^{*,s}$ with high probability, but this is beyond the scope of our current work.  

As obvious as the form of the penalty term in \eqref{eqn:penterm-aicc}
may seems in retrospect, i.e., not counting the zero entries as a part of parameters, to our best knowledge, surprisingly, there is no literature that addresses this idea as we did. The idea of not counting the zeros is similar to McNemar's test in the way it is discussed in \cite[pg 77]{brian1996pattern}, although the similarity is only in sprit.
On the other hand, as done in \cite{bouguila2008clustering}, often clustering of multinomial observations is studied in a Bayesian manner, where the prior distribution on a probability vector is specified by a
non-degenerate Dirichlet distribution.  For such situations, as discussed, our criterion with $(s,\gamma)=(1,1)$ reduces to the conventional AIC.  

Beyond data of biological nature akin to our numerical section, 
our work in this paper can also be considered for various types of
data with noise or without noise,  which can be expected to be decomposed as a product of two non-negative matrices.  For example, a collection of images of single color channel, a collection of documents with various topics and a collection of sensors interaction  records are few such examples.    A further investigation into how far our approach can be taken to determine the non-negative factorization's inner dimension, is of interest beyond this work.

\appendices

\section{Proof of Theorem \ref{thm:result/main}}
\begin{proof}
First, by way of a Taylor expansion of the $\log$ function, we note
\begin{align}
&\quad\mathbf E[\varphi(\widehat P)    ] \\
&= \varphi( P^*) 
- \mathbf E[ \sum_{i,t} \frac{1}{ \overline n_{\kappa^*(t)}^2} P^*_{i,t} \bm
1\{ P^*_{i,t} > 0\} \frac{1}{ P^*_{i,t}} (\widehat P_{i,t} -
 P^*_{i,t})    ]\label{eqn:biascorrection-1} \\
& - \mathbf E[ \sum_{i,t}  \frac{1}{ \overline n_{\kappa^*(t)}^2}   P^*_{i,t} \bm 1\{ P^*_{i,t} > 0\} \frac{-1}{2 (P^*_{i,t})^2}
(\widehat P_{i,t} -  P^*_{i,t})^2   ] \label{eqn:biascorrection-1.5}\\
& -  \mathbf E[ \sum_{i,t}  \frac{1}{ \overline n_{\kappa^*(t)}^2}  R(\widehat P_{i,t},  P^*_{i, t}) ],\label{eqn:biascorrection-2}
\end{align}   
where $R$ denotes the high order remainder term.

Since $\widehat P$ is an unbiased estimator of $ P^*$, we see that the second
term on the right in \eqref{eqn:biascorrection-1} vanishes to zero.  For the
 term in \eqref{eqn:biascorrection-1.5}, we note that since each $X_{i,t}$
is a binomial random variable for  $N_t$ trials with its success probability
$ P^*_{i,t}$, we see that
\begin{align}
    &\quad -\sum_{i,t} \frac{1}{ \overline n_{\kappa^*(t)}^2}     \bm 1\{ P^*_{i,t} > 0\} \frac{-1}{2 P^*_{i,t}} \mathbf E[(\widehat P_{i,t} -  P^*_{i,t})^2 ] \\
    & = \sum_{i,t}  \frac{1}{ \overline n_{\kappa^*(t)}^2}  \bm 1\{ P^*_{i,t} > 0\} \frac{1}{2 P^*_{i,t}} \frac{1}{N_t}  P^*_{i,t} (1-  P^*_{i,t}) \\
    & = \sum_{i,t}  \frac{1}{ \overline n_{\kappa^*(t)}^2}  \bm 1\{ P^*_{i,t} > 0\} \frac{1}{2N_t} (1-  P^*_{i,t}) \\
    & = \sum_{t=1}^T \frac{1}{ \overline n_{\kappa^*(t)}^2}  \frac{1}{2N_t} \left(\sum_{i} \bm 1\{ P^*_{i,t} >0\}\right) \\ 
    &\qquad\qquad - \sum_{t=1}^T   \frac{1}{ \overline n_{\kappa^*(t)}^2}  \frac{1}{2N_t} \left(\sum_{i} \bm 1\{ P^*_{i,t} > 0\}  P^*_{i,t}\right) \\
    & =  \sum_{t=1}^T \frac{1}{ \overline n_{\kappa^*(t)}^2}  \frac{\overline Z_{\kappa^{*}(t)}}{2N_t} - \sum_{t=1}^T  \frac{1}{ \overline n_{\kappa^*(t)}^2}  \frac{1}{2N_t},\label{eqn:biascorrection-3}
\end{align}
where the last equality is due to the fact 
that each column of $ P^*$ sums to one.  Hence, in summary, we 
see that 
\begin{align}
\lim_{\ell\rightarrow\infty} 
\ell (\mathbf E[\varphi(\widehat P)  ] - \varphi( P^*))
= \lim_{\ell\rightarrow\infty} \ell\sum_{t=1}^T \frac{\overline Z_{\kappa^{*}(t)}-1}{2N_t { \overline n_{\kappa^*(t)}^2} }. \label{eqn:AICc/proof/part/one}
\end{align}

Letting $t_{k}$ be any fixed $t \in k(\kappa^{*})$, since $\lim_{\ell\rightarrow\infty} N_{t_k}/\ell =   \overline \lambda_k$ by assumption, 
\begin{align}
\lim_{\ell\rightarrow\infty} \ell \sum_{t=1}^T \frac{\overline Z_{\kappa^{*}(t)}-1}{N_t \overline n_{\kappa^*(t)}^2} 
&= \lim_{\ell\rightarrow\infty} \ell \sum_{k=1}^{K^{*}} \overline n_k \frac{ \overline Z_{k}-1}{ N_{t_k} \overline n_{k}^2}  \\
&=   \sum_{k=1}^{K^{*}}  \frac{ \overline Z_{k} -1}{ \overline n_{k} \lim_{\ell\rightarrow\infty}  N_{t_k}/\ell } \\
&=  \sum_{k=1}^{K^{*}}  \frac{ \overline Z_{k}-1}{\overline \lambda_k \overline n_k}.
    \label{eqn:AICc/proof/part/two}
\end{align}

We next consider $\mathbf E[ \sum_{i,t} R(\widehat P_{i,t},  P^*_{i,t})]$.
Note that
\begin{align}
&\quad R(\widehat P_{i,t},  P_{i,t}^*) \\
& = \sum_{k=3}^{\infty} \frac{1}{k}\frac{(-1)^{k+1}}{ (P^*_{i,t})^{k}}|\widehat P_{i,t} -  P^*_{i,t}|^{k} \\
& = \frac{-1}{ (P^*_{i,t})^{3}} (\widehat P_{i,t} -  P^*_{i,t})^{3} 
\sum_{k=0}^{\infty} \frac{(-1)^{k+1}}{ (P^*_{i,t})^{k}} \frac{(\widehat P_{i,t} -  P^*_{i,t})^{k}}{k+3}.
\end{align}
Hence, 
\begin{align}
&\ell |R(\widehat P_{i,t},  P_{i,t}^*)|  \\
&\le  \frac{\ell}{ (P^*_{i,t})^{3}} |\widehat P_{i,t} -  P^*_{i,t}|^{3} 
\sum_{k=0}^{\infty} \frac{1}{ (P^*_{i,t})^{k}} \frac{|\widehat P_{i,t} -  P^*_{i,t}|^{k}}{k}.
\end{align}
Since $\widehat P_{i,t} \rightarrow  P^*_{i,t}$ almost surely,    
it can be shown that there exists a constant $c > 0$ such that for each sufficiently small $\varepsilon > 0$, 
for sufficiently large $\ell$,
with $1-\varepsilon$ probability,
$$
\sum_{k=0}^{\infty} \frac{1}{ (P^*_{i,t})^{k}} \frac{|\widehat P_{i,t} -  P^*_{i,t}|^{k}}{k} \le c.
$$
Moreover, using the third moment formula for a binomial random variable explicitly, we have 
\begin{align}
&\qquad \lim_{\ell\rightarrow\infty}\ell \mathbf E[ |\widehat P_{i,t} -  P^*_{i,t}|^{3}\left| \bm N \right.  ] \\
&\le 
\lim_{\ell\rightarrow\infty} \ell \frac{1}{N_{t}^{3}}  N_{t}
 P^*_{i,t} (1- P^*_{i,t}) (1- 2  P^*_{i,t}) \\
&\le 
\lim_{\ell\rightarrow\infty} \frac{1}{N_{t}/\ell}
 P^*_{i,t} (1- P^*_{i,t}) 
\lim_{\ell\rightarrow\infty}   \frac{1- 2  P^*_{i,t}}{N_{t}} = 0.
\end{align}

In summary, $\lim_{\ell\rightarrow\infty }\ell\mathbf E[R(\widehat P_{i,t},  P_{i,t}^*)] = 0$.  
Combining with \eqref{eqn:AICc/proof/part/one} and \eqref{eqn:AICc/proof/part/two}, this completes our proof. 
\end{proof}

\section{Proof of Theorem \ref{thm:main:ModelSelectThm}}

We first focus on the under-fitted case. That is, consider
$\Delta(K^*) - \Delta(K^* - 1)$. First,   for $(\kappa^*, Q^*) \in \Theta(K^*)$, we have
$\log(f(X;\kappa^*,Q^*))
	= \sum_{t=1}^T \log \binom{N_t}{X_t} 
	+ \sum_{k=1}^{K^*} \sum_{i=1}^d \sum_{\{t: \kappa^*(t) = k\}} X_{it} \log Q^*_{ik}$, 
and similarly, but specializing for merging of the $K^* - 1$ and $K^*$ blocks from the true parameter structure, for $(\kappa, Q) \in \Theta_{K^*-1}$ with merging of the $(K^{*}-1)$st and $K^{*}$th clusters,
\begin{align*}
	& \log(f(X;\kappa,Q)) \\
	= & \sum_{t=1}^T \log \binom{N_t}{X_t} + \sum_{k=1}^{K^* - 2} \sum_{i=1}^d \sum_{\{t: \kappa(t) = k\}} X_{it} \log Q_{ik} \\ 
    & + \sum_{i=1}^d \sum_{\{t: \kappa(t) = K^* - 1\}} X_{it} \log Q_{i, K^* - 1} \\
	= & \sum_{t=1}^T \log \binom{N_t}{X_t} + \sum_{k=1}^{K^* - 2} \sum_{i=1}^d \sum_{\{t: \kappa^*(t) = k\}} X_{it} \log Q^*_{ik} \\ 
    & + \sum_{k=K^* - 1}^{K^*} \sum_{i=1}^d \sum_{\{t: \kappa^*(t) = k\}} X_{it} \log Q_{i,K^*-1}.
\end{align*}

Hence, it follows that
\begin{align*}
	& \log(f(X;\kappa^*,Q^*)) - \log(f(X;\kappa,Q)) \\
    = & \sum_{k=K^* - 1}^{K^*} \sum_{i=1}^d \sum_{\{t: \kappa^*(t) = k\}} X_{it} \log (Q^*_{ik}/Q_{i,K^*-1}).
\end{align*}
Then, by taking the expectation with respect to the probability mass function defined
by $f(\cdot|\kappa^*, Q^*)$, define
\begin{align*}
	\delta^{*,m}(T) := & \sum_{k=K^* - 1}^{K^*} \sum_{i=1}^d \sum_{\{t: \kappa^*(t) = k\}} \frac{\mathbf E[X_{it}]}{N_{t}}\log (Q^*_{ik}/Q_{i,K^*-1}) \\
    = & \sum_{k=K^* - 1}^{K^*} \left(n_k(\kappa^*) \sum_{i=1}^d  Q^*_{ik} \log (Q^*_{ik}/Q_{i,K^*-1})\right),
\end{align*}
where $n_k(\kappa^*)  = \sum_{t=1}^T \bm 1\{ \kappa^*(t) = k \} = \overline n_k$. 

Next, note that 
\begin{align*}
	&\Delta^{*,m}(K^* - 1) - \Delta^{*}(K^*) \\
	= & \sum_{k=K^* - 1}^{K^*} \sum_{i=1}^d \sum_{\{t: \kappa^*(t) = k\}} \frac{X_{it}}{N_{t}} \log (Q^*_{ik}/Q_{i,K^*-1})\\
    & + \texttt{penalty}^{*,m}(K^*-1) - \texttt{penalty}^*(K^*).
\end{align*}
Hence, we have
\begin{align*}
	& \Delta^{*,m}(K^* - 1) - \Delta^{*}(K^*) \\
	= & \sum_{k=K^* - 1}^{K^*} \sum_{i=1}^d \sum_{\{t: \kappa^*(t) = k\}} \frac{X_{it}}{N_t} \log (Q^*_{ik}/Q_{i,K^*-1}) - \delta^{*,m}(T)\\
    & + \texttt{penalty}^{*,m}(K^*-1) - \texttt{penalty}^{*}(K^*) + \delta^{*,m}(T) \\
    = & \sum_{k=K^* - 1}^{K^*} \sum_{i=1}^d \sum_{\{t: \kappa^*(t) = k\}}\left(\frac{X_{it}}{N_t} - \mathbf E[\frac{X_{it}}{N_t} ]\right) \log\left(\frac{Q^*_{ik}}{Q_{i,K^*-1}}\right) \\
    & + \texttt{penalty}^{*,m}(K^*-1) - \texttt{penalty}^{*}(K^*) + \delta^{*,m}(T).
\end{align*}

Let 
\begin{align*}
\zeta^{*,m} \equiv \sum_{k=K^* - 1}^{K^*} \sum_{i=1}^d \sum_{\{t:\in k(\kappa^*)\}} \left(\frac{X_{it}}{N_t} - \mathbf E[\frac{X_{it}}{N_t} ]\right)\log \left(\frac{Q^*_{ik}}{Q_{i,K^*-1}}\right)
\end{align*}
and note that almost surely,  $\lim_{\ell\rightarrow\infty} \zeta = 0$.
Now, since
\begin{align*}
& \lim_{\ell \rightarrow \infty} \mathbf P[{\zeta}^{*,m}  +  (\texttt{penalty}^{*,m}  - \texttt{penalty}^{*} + \delta^{*,m}) > 0] \\
    = 	& \lim_{\ell \rightarrow \infty} \mathbf P[(\Delta^{*,m} - \Delta^{*}) > 0] \\
    =	& \lim_{\ell \rightarrow \infty} \mathbf P[\Delta^{*,m}    > \Delta^{*} ],
\end{align*}
to show our claim, it is enough to show that for sufficiently large $\ell$, 
\begin{align*}
\texttt{penalty}^{*,m} - \texttt{penalty}^{*} + {\delta^{*,m}} > 0.
\end{align*} 
In general, for $(s,\gamma) = (1/2, \log(N))$, as $N\rightarrow\infty$, 
\begin{align*}
&\quad \texttt{penalty}(K;s,\gamma)  \\
&=   \log(N) \sum_{k=1}^{K} \frac{\widehat Z_{k}^{{(K)}}-1}{\sqrt{\widehat N_{k}^{(K)}}} \\
&=   \frac{\log(N)}{\sqrt{N}} \sum_{k=1}^{K}
\left( \frac{1}{\sqrt{\widehat N_{k}^{(K)}/N}} (\widehat Z_{k}^{{(K)}}-1)\right) 
\rightarrow 0,
\end{align*}
whence
\begin{align*}
&\lim_{N\rightarrow\infty} \left(\texttt{penalty}^{*,m}   -  \texttt{penalty}^{*} + {\delta^{*,m}}\right)    \\
=   &  \sum_{k=K^{*}-1}^{K^*} \sum_{i=1}^d n_{k}(\kappa^{*})Q^{*}_{ik} \log (Q^*_{ik}/Q_{i,K^*-1}^{*,m}) > 0,
\end{align*}
as desired.  This completes the under-fitting case.

For the over-fitting case,  let 
\begin{align*}
	\delta^{*,s} := & \sum_{k=K^*}^{K^*+1} \sum_{i=1}^d \sum_{\{t: \kappa^*(t) = k\}} \frac{E[X_{it}]}{N_{t}}\log (Q_{i,K^*}^{*}/Q_{ik}) \\
   & =  \sum_{k=K^*}^{K^*+1} \left(n_k(\kappa^*) \sum_{i=1}^d  Q^*_{iK^{*}} \log (Q_{i,K^*}^{*}/Q_{ik})\right),
\end{align*}
where $n_k(\kappa^*)  = \sum_{t=1}^T \bm 1\{ \kappa^*(t) = k \} = \overline n_k$, and let
\[
\zeta^{*,s} \equiv \sum_{k=K^*}^{K^*+1} \sum_{i=1}^d {\sqrt{N}} \sum_{t \in k(\kappa^{*,s})}\left(\frac{X_{it}}{N_t} - \mathbf E[\frac{X_{it}}{N_t} ]\right) \log (\frac{Q^*_{i,K^*}}{Q_{i,k}^{*,s}}).
\]
Since $Q_{i,k} = Q_{i,K^{*}-1}^{*}$ for each $k \ge K^{*}-1$, we have 
\begin{align*}
\delta^{*,s} \equiv 0 \text{ and } \zeta^{*,s} \equiv 0.
\end{align*}
Then, we have
\begin{align*}
& \lim_{N \rightarrow \infty} \mathbf P[\Delta^{*,s}    > \Delta^{*} ] \\
&  =  \lim_{N \rightarrow \infty} \mathbf P[\sqrt{N}{\zeta}^{*,s}  + \sqrt{N} (\texttt{penalty}^{*,s}  - \texttt{penalty}^{*}  +  \delta^{*,s} ) > 0]\\
&  = \lim_{N \rightarrow \infty} \mathbf P[\sqrt{N} (\texttt{penalty}^{*,s}  - \texttt{penalty}^{*}) > 0].
\end{align*}
Now, 
\begin{align*}
\qquad & \texttt{penalty}^{*,s} -  \texttt{penalty}^{*} \\
& =  
\log(N) \sum_{k=1}^{K^{*}+1} \frac{\overline Z_{k}(Q^{*,s})-1}{\sqrt{N_{k}^{*,s}}}
- \log(N) \sum_{k=1}^{K^{*}} \frac{\overline Z_{k}(Q^{*})-1}{\sqrt{N_{k}^{*}}} \\ 
&\ge  \log(N) \frac{\overline Z_{K^{*}+1}(Q^{*,s})-1}{\sqrt{N_{K^{*}+1}^{*,s}}} > 0,
\end{align*}
where  $\overline Z_{k}(Q^{*,s})$ counts the number of non-zero entries of $Q_{k}^{*,s}$, $N_k^* = (\sum_{t \in k(\kappa^*)} N_t)^{1/2}$,

and $N_k^{*,s} = (\sum_{t \in k(\kappa^{*,s})} N_t)^{1/2}$.
Hence, 
\begin{align*}
& \lim_{N\rightarrow\infty} \sqrt{N} (\texttt{penalty}^{*,s} -  \texttt{penalty}^{*}) \\
 & \ge  \lim_{N\rightarrow\infty} \log(N) \frac{\overline Z_{K^{*}+1}(Q^{*,s})-1}{\sqrt{N_{K^{*}+1}^{*,s}/N}} \\
 & \ge    \lim_{N \rightarrow\infty}  \log(N) \frac{\overline Z_{K^{*}+1}(Q^{*,s})-1}{\sqrt{ \lim_{N \rightarrow\infty}N_{K^{*}+1}^{*,s}/N  }}  = \infty.
\end{align*} 
Hence, it follows that, as desired, 
\begin{align*}
&\quad \lim_{\ell \rightarrow \infty} \mathbf P[\Delta^{*,s}    > \Delta^{*} ]  \\
&=  \mathbf P[ \lim_{N\rightarrow\infty} \sqrt{N} (\texttt{penalty}^{*,s} -  \texttt{penalty}^{*})  > 0]  = 1.
\end{align*}
This completes our proof.  

As a side note, we finish by observing that the last part of our argument can be slightly generalized.  Specifically, 
by the central limit theorem (CLT), as $N \rightarrow \infty$,
\[
\sum_{\{t:\kappa(t)=k\}}
\frac{ (X_{it}/N_t -  Q^*_{ik})}
    {\sqrt{Q^*_{ik}(1 - Q^*_{ik})/N_{t}}} 
	\implies
    \mathcal{N}(0, 1),
\]
where the convergence is in distribution, and similarly, we have as $N \rightarrow \infty$,
\[
\sum_{\{t:\kappa(t)=k\}}
    \sqrt{N_{t}} { (X_{it}/N_t-  Q^*_{ik})}
    \implies
    \mathcal{N}(0, Q^*_{ik}(1 - Q^*_{ik})),
\]
where $\mathcal{N}(0, Q^*_{ik}(1 - Q^*_{ik}))$ denotes a normal random variable with mean zero and variance $Q^*_{ik}(1 - Q^*_{ik})$. 
In fact, by the multivariate CLT, we have that as $N \rightarrow \infty$, $\zeta^{*,s}$ converges to a linear combination of normal  random variables with mean zero even when $Q_{K^*}^{*,s}$ and  $Q_{K^*}^{*,s}$ may not equal the value of $Q_{K^*}^{*}$.  This allows one to extend the last part of the over-fitting case.  
\hfill\qed

\section{Proof of Theorem  \ref{thm:meantotalinteraction}}

We first start with the following lemma.
\begin{my.lemma}\label{thm:errorbound}
For each $n$, $T$ and $C$, 
\begin{align}
\quad & \frac{1}{\sqrt{dT}} \mathbf E[\|\widehat Y - \mathbf E[X]\|_F] \\
\le &\  C \sqrt{r}
\left(\gamma_{1}  \left(\frac{ {d}}{T^{3}}\right)^{1/4} +
\gamma_2 
\frac{\left(\log(T)\right)^{1/2}}{ T}
\right)
\\
&\quad +
\sqrt{
\frac{1}{dT}
\sum_{i,t} \left( \mathbf E[(X_{i,t} - C)^+]\right)^2 },\label{eqn:errorbound}
\end{align}
where  $\gamma_1,\gamma_2,\gamma_3$
are (universal) constants that do not depend on $C$, $n$ and $T$ and 
$(X_{i,t}-C)^+ = \max\{X_{i,t}-C,0\}$.
\end{my.lemma}
\begin{proof}
Let
\begin{align*}
\MSE(\widehat Y)  :=  \frac{1}{dT} \sum_{i,t} (\widehat Y_{i,t} - \mathbf E[Y_{i,t}])^2.
\end{align*}
By a triangular inequality, we have 
\begin{align*}
& \frac{1}{\sqrt{dT}}
\mathbf E[\|\widehat Y - \mathbf E[X]\|_F]\\
\le &
\frac{1}{\sqrt{dT}}
\mathbf E[\|\widehat Y - \mathbf E[Y]\|_F]
+ 
\frac{1}{\sqrt{dT}}
\|\mathbf E[Y] - \mathbf E[X]\|_F \\
\le &
\sqrt{\MSE(\widehat Y)} + 
\sqrt{
\frac{1}{dT}
\|\mathbf E[Y] - \mathbf E[X]\|_F^2}\\
\le &
\sqrt{C^2 \left(\frac{1}{C^2}\MSE(\widehat Y)\right)}+ 
\sqrt{
\frac{1}{dT}
\|\mathbf E[Y] - \mathbf E[X]\|_F^2}.
\end{align*}
Now, by \cite[Theorem 1.1 \& 1.3]{kmo10noise}, for  
some fixed (universal) constant $\gamma_1,\gamma_2$
(in particular, not depending on $C$, $n$ and $T$),  we have
\begin{align}
\frac{1}{C^2}\MSE(\widehat Y) 
&\le 
\gamma_{1}  \sqrt{r} \left(\frac{ {d}}{T^{3}}\right)^{1/4} \\
&\qquad +
\gamma_2 
\frac{\sqrt{r}}{ d^{1/2} T^{1/2}}
\left(\frac{d \log(T)}{T}\right)^{1/2}.
\end{align}
On the other hand, 
\begin{align*}
\|\mathbf E[X - Y]\|_F^2
& = \sum_{i,t} \left( \mathbf E[X_{i,t} - Y_{i,t};X_{i,t} > C]\right)^2 \\
& =  \sum_{i,t} \left( \mathbf E[(X_{i,t} - C)^+]\right)^2 ,
\end{align*}
where in the second equality, we have used 
the fact that on the event $\{ X_{ij,t} \le C\}$, 
we have $X_{ij,t} = Y_{ij,t}$ for all $ij$ and $t$.  
Our claim follows from this. 
\end{proof}

\begin{proof}[Proof of Theorem \ref{thm:meantotalinteraction}]
By assumption, we have 
\begin{align*}
0 = \lim_{d\wedge T\rightarrow\infty}
C_{d,T}  \left(\gamma_{1}  \left(\frac{ {d}}{T^{3}}\right)^{1/4} +
\gamma_2 
\frac{\left(\log(T)\right)^{1/2}}{ T}
\right).
\end{align*}
Next, to complete our proof, it is enough to show that 
$\lim_{d\wedge T \rightarrow\infty} e(d,T) = 0$, where
\begin{align*}
e(d,T) :=
\sqrt{
\frac{1}{dT}
\sum_{i,t} \left( \mathbf E[(X_{i,t} - C_{d,T})^+]\right)^2 }.
\end{align*}
Note that for sufficiently large values of $d \wedge T$,  
$C_{d,T} \ge (\mathbf E[X_{i,t}])$, and we have 
\begin{align*}
&\qquad  \mathbf E[(X_{i,t}-C_{d,T})^+] \\
&=
\sum_{m=C_{d,T}+1}^\infty (m-C_{d,T}) \frac{(\mathbf E[X_{i,t}])^m}{m!} \exp(-(\mathbf E[X_{i,t}]))\\
&=
(\mathbf E[X_{i,t}])
\sum_{m=C_{d,T}+1}^\infty \frac{(\mathbf E[X_{i,t}])^{m-1}}{(m-1)!} \exp(-(\mathbf E[X_{i,t}])) \\
&\qquad\qquad -
C_{d,T} \sum_{m=C_{d,T}+1}^\infty \frac{(\mathbf E[X_{i,t}])^m}{m!} \exp(-(\mathbf E[X_{i,t}]))\\
&\le
C_{d,T}
\sum_{m=C_{d,T}}^\infty \frac{(\mathbf E[X_{i,t}])^m}{m!} \exp(-(\mathbf E[X_{i,t}])) \\
&\qquad\qquad -
C_{d,T} \sum_{m=C_{d,T}+1}^\infty \frac{(\mathbf E[X_{i,t}])^m}{m!} \exp(-(\mathbf E[X_{i,t}]))\\
&\le C_{d,T} \frac{(\mathbf E[X_{i,t}])^{C_{d,T}}}{C_{d,T}!} \exp(-(\mathbf E[X_{i,t}])) \\
& = (\mathbf E[X_{i,t}]) \frac{(\mathbf E[X_{i,t}])^{C_{d,T}-1}}{(C_{d,T}-1)!} \exp(-(\mathbf E[X_{i,t}])).
\end{align*}
Then,
\begin{align*}
&\frac{1}{|\mathcal B|}\sum_{i,t} (\mathbf E[(X_{i,t} - C_{d,T})^+])^2 \\
&\le 
\frac{1}{|\mathcal B|} \sum_{i,t} (\mathbf E[X_{i,t}])^2 \exp(-2(\mathbf E[X_{i,t}]))   \left(\frac{(\mathbf E[X_{i,t}])^{C_{d,T}-1}}{(C_{d,T}-1)!}\right)^2\\
&=
\frac{1}{|\mathcal B|} \sum_{b=1}^B |\mathcal B_b|
\nu_b^2 \exp(-2\nu_b)  \left(\frac{\nu_{b}^{C_{d,T}-1}}{(C_{d,T}-1)!}\right)^2\\
&=
\sum_{b=1}^B \frac{|\mathcal B_b|}{|\mathcal B|}  \nu_b^2 \exp(-2\nu_b)  \left(\frac{\nu_{b}^{C_{d,T}-1}}{(C_{d,T}-1)!}\right)^2.
\end{align*}
Then, $\lim_{d\wedge T\rightarrow\infty} e(n,T) = 0$ since
\begin{align*}
&0 \le \limsup_{d\wedge T \rightarrow \infty} 
\frac{1}{|\mathcal B|}\sum_{i,t} (\mathbf E[(X_{i,t} - C_{d,T})^+])^2 \\
&= \sum_{b=1}^B p_b  
\limsup_{d\wedge T \rightarrow \infty}
\left(\nu_b^2 \exp(-2\nu_b)  
\left(\frac{\nu_{b}^{C_{d,T}-1}}{(C_{d,T}-1)!}\right)^2\right) = 0.
\end{align*}
\end{proof}

\section{Derivation of an objective function for computing $\MLqE$ via a Markov chain Monte Carlo method}\label{sec:MCMC-MLqE-objective}
Fix $q < 1$ and $\kappa : \{1,\ldots, T\} \rightarrow \{1,\ldots, K\}$.  We let $\mathcal Q$ to be the free variables.
Now, given $X_{1}, X_{2},\ldots, X_{T}$, we let 
\begin{align*}
L(\kappa,\mathcal Q)  
& =  \frac{1}{1-q} \sum_{t=1}^{T} \sum_{i=1}^{d} X_{i}(t) Q_{i,\kappa(t)}^{1-q},\\
& =  \frac{1}{1-q} \sum_{k=1}^{K} \sum_{i=1}^{d} \left(\sum_{\{t: \kappa(t) = k \}}X_{i}(t)\right) Q_{i,k}^{1-q},
\end{align*}
where for each $k=1,\ldots, K$, $\sum_{i=1}^{d} Q_{i,k} = 1$.  
Then, formulating the problem using the first order KKT  condition reduces the problem of maximizing $L(\kappa,\mathcal Q)$ with respect to $\mathcal Q$ to maximizing 
\begin{align*}
\frac{1}{1-q} \sum_{k=1}^{K} \sum_{i=1}^{d} M_{i,k} Q_{i,k}^{1-q} + 
\sum_{k=1}^{K} \mu_{k} \left(1- \sum_{i=1}^{d}Q_{i,k}\right),
\end{align*} 
where $\mu_{1}, \ldots, \mu_{K}$ denote the Lagrange multipliers and we write
$M_{i,k} = \left(\sum_{\{t: \kappa(t) = k \}}X_{i}(t)\right)$.

Specifically, taking a derivative with respect to 
each $Q_{k,i}$ yields the condition that for each $k, i$,  
\begin{align*}
0 = M_{i,k} Q_{i,k}^{-q} + \mu_{k} (-1),
\end{align*}
whence we have $\mu_{k}^{1/q}Q_{i,k} = M_{i,k}^{1/q} $, $\mu_{k}^{1/q} = \sum_{i=1}^{d} M_{i,k}^{1/q}$ and from this observation, 
we set
\begin{align*}
\widehat Q_{i,k} := \frac{M_{i,k}^{1/q}}{\sum_{i=1}^{d} M_{i,k}^{1/q}}.
\end{align*}
Now, let 
\begin{align}
L^{*}(\kappa;q) =   \sum_{k=1}^{K} \sum_{i=1}^{d}  M_{i,k} \left(\frac{M_{i,k}^{1/q}}{\sum_{j=1}^{d} M_{k,j}^{1/q}}\right)^{1-q}, \label{eqn:MLqE-explicit}
\end{align}
where the dependence of $M_{i,k}$ on $\kappa$ is implicitly stated.  Also, performing a similar sequence of computations, for $q=1$, i.e., for the maximum likelihood estimator, we have
\begin{align}
L^{*}(\kappa;q) =   \sum_{k=1}^{K} \sum_{i=1}^{d}  M_{i,k} \log\left(\frac{M_{i,k}}{\sum_{j=1}^{d} M_{k,j}}\right). \label{eqn:MLE-explicit}
\end{align}

Then, taking $\kappa$ as a free variable,  the values of $L^{*}(\kappa;q)$ can be explored over all $\kappa \in \{1,\ldots, K\}^{T}$ using any Markov chain Monte Carlo method, for example, by a Gibbs sampling approach.   We leave to the reader the remaining details for a Gibbs sampling procedure in which at each step, a single coordinate of $\kappa$ can be changed.

\section{Connection to the conventional AIC and BIC}\label{sec:AICBICtoPen}
In this section, we show that the form of the penalty term 
in \eqref{eqn:penterm-aicc} reduces to the conventional AIC and BIC criteria under some simplifying conditions.  

Specifically, we assume in this section
that $N_{0} = N_{1} = N_{2} = \cdots = N_{T} > 0$ and for each $K$, 
that $\widehat Q_{ik}^{(K)} > 0$ for all $i=1,\ldots, d$ and $k=1,\ldots, K$.  

First, to see a connection to the conventional AIC criterion in 
\eqref{eqn:DeltaAIC-classic}, we note that 
\begin{align*}
& N_{0} \Delta(K)  \\
& = -  \sum_{t=1}^{T}\sum_{i=1}^{d} X_{it} \log(\widehat Q_{i\widehat\kappa(t)}^{(K)})  +  N_{0}\texttt{penalty}(K;1,1),
\end{align*}
where 
\begin{align*}
N_{0}\texttt{penalty}(K;1,1) = \sum_{k=1}^{K} \left(\widehat Z_{k}^{(K)} - 1\right) = dK - K.
\end{align*}
Hence, for some constant $C$ that depends on only the value of $X$,  
\begin{align*}
N_{0} \Delta(K) = C  
-\log(f_{X}(X | \widehat \theta(K))) + \texttt{penalty}_{\AIC}(K),
\end{align*}
which differs from the one in \eqref{eqn:DeltaAIC-classic} by the additive constant $C$. 

Next, to see a connection to the conventional BIC criterion in 
\eqref{eqn:DeltaBIC-classic}, we note that when $\gamma = \log(N)/\sqrt{N_{0}}$ 
and $s=1/2$, 
$$
N_{0}\Delta(K) 
= -\log(f_{X}(X | \widehat \theta(K))) 
+ N_{0} \texttt{penalty}(K;\gamma,1/2).
$$
Then, we have 
$$
N_{0}\texttt{penalty}(K;\gamma,1/2) = \log(N)  (dK - K).
$$
Hence,  
\begin{align*}
N_{0}\Delta(K) = C  -\log(f_{X}(X | \widehat \theta(K))) + \texttt{penalty}_{\BIC}(K),
\end{align*}
which differs from the one in \eqref{eqn:DeltaBIC-classic} by the additive constant $C$. 

\subsection*{Acknowledgment}
{This work is partially supported by Johns Hopkins University
Armstrong Institute for Patient Safety and Quality and
the XDATA program of the Defense Advanced Research
Projects Agency (DARPA) administered through
Air Force Research Laboratory contract FA8750-12-2-0303. We thank Youngser Park for his assistance in performing numerical experiments. We thank the anonymous referees for their valuable comments.}

\bibliographystyle{IEEEtran}

\end{document}